# Effect of Depth and Width on Local Minima in Deep Learning

**Kenji Kawaguchi**
*kawaguch@mit.edu*
*MIT, Cambridge, MA 02139, U.S.A.*

**Jiaoyang Huang**
*jiaoyang@math.harvard.edu*
*Harvard University, Cambridge, MA 02138, U.S.A.*

**Leslie Pack Kaelbling**
*lpk@csail.mit.edu*
*MIT, Cambridge, MA 02139, U.S.A.*

**In this paper, we analyze the effects of depth and width on the quality of local minima, without strong overparameterization and simplification assumptions in the literature. Without any simplification assumption, for deep nonlinear neural networks with the squared loss, we theoretically show that the quality of local minima tends to improve toward the global minimum value as depth and width increase. Furthermore, with a locally induced structure on deep nonlinear neural networks, the values of local minima of neural networks are theoretically proven to be no worse than the globally optimal values of corresponding classical machine learning models. We empirically support our theoretical observation with a synthetic data set, as well as MNIST, CIFAR-10, and SVHN data sets. When compared to previous studies with strong overparameterization assumptions, the results in this letter do not require overparameterization and instead show the gradual effects of overparameterization as consequences of general results.**

## 1 Introduction

Deep learning with neural networks has been a significant practical success in many fields, including computer vision, machine learning, and artificial intelligence. Along with its practical success, deep learning has been theoretically analyzed and shown to be attractive in terms of its expressive power. For example, neural networks with one hidden layer can approximate any continuous function (Leshno, Lin, Pinkus, & Schocken, 1993; Barron, 1993), and deeper neural networks enable us to approximate functions of certain classes with fewer parameters (Montufar, Pascanu, Cho, & Bengio, 2014; Livni, Shalev-Shwartz, & Shamir, 2014; Telgarsky, 2016).

However, training deep learning models requires us to work with a seemingly intractable problem: nonconvex and high-dimensional optimization. Finding a global minimum of a general nonconvex function is NP-hard (Murty & Kabadi, 1987), and nonconvex optimization to train certain types of neural networks is also known to be NP-hard (Blum & Rivest, 1992). These hardness results pose a serious concern only for high-dimensional problems, because global optimization methods can efficiently approximate global minima without convexity in relatively low-dimensional problems (Kawaguchi, Kaelbling, & Lozano-Pérez, 2015).

A hope is that beyond the worst-case scenarios, practical deep learning allows some additional structure or assumption to make nonconvex high-dimensional optimization tractable. Recently, it has been shown with strong simplification assumptions that there are novel loss landscape structures in deep learning optimization that may play a role in making the optimization tractable (Dauphin et al., 2014; Choromanska, Henaff, Mathieu, Ben Arous, & LeCun, 2015; Kawaguchi, 2016). Another key observation is that if a neural network is strongly overparameterized so that it can memorize any data set of a fixed size, then all stationary points (including all local minima and saddle points) become global minima, with some nondegeneracy assumptions. This observation was explained by Livni et al. (2014) and further refined by Nguyen and Hein (2017, 2018). However, these previous results (Livni et al., 2014; Nguyen and Hein, 2017, 2018) require strong overparameterization by assuming not only that a network's width is larger than the data set size but also that optimizing only a single layer (the last layer or some hidden layer) can memorize any data set based on an assumed condition on the rank or nondegeneracy of other layers.

In this letter, we analyze the effects of depth and width on the values of local minima, without the strong overparameterization and simplification assumptions in the literature. As a result, we prove quantitative upper bounds on the quality of local minima, which shows that the values of local minima of neural networks are guaranteed to be no worse than the globally optimal values of corresponding classical machine learning models, and the guarantee can improve as depth and width increase.

## 2 Preliminaries

This section defines the optimization problem considered in this letter and introduces the basic notation.

**2.1 Problem Formulation.** Let $x \in \mathbb{R}^{d_x}$ and $y \in \mathbb{R}^{d_y}$ be an input vector and a target vector, respectively. Let $\{(x_i, y_i)\}_{i=1}^{m}$ be a training data set of size $m$. Given a set of $n$ matrices or vectors $\{M^{(j)}\}_{j=1}^{n}$, define $[M^{(j)}]_{j=1}^{n} :=$ $\begin{bmatrix} M^{(1)} & M^{(2)} & \cdots & M^{(n)} \end{bmatrix}$ to be a block matrix of each column block being $M^{(1)}, M^{(2)}, \ldots, M^{(n)}$. Define the training data matrices as $X := ([x_i]_{i=1}^m)^\top \in \mathbb{R}^{m \times d_x}$ and $Y := ([y_i]_{i=1}^m)^\top \in \mathbb{R}^{m \times d_y}$.

This letter considers the squared loss function, with which the training objective of the neural networks can be formulated as the following optimization problem:

$$\underset{\theta}{\text{minimize}} \; L(\theta) := \frac{1}{2} \|\hat{Y}(X, \theta) - Y\|_F^2, \tag{2.1}$$

where $\|\cdot\|_F$ is the Frobenius norm, $\hat{Y}(X, \theta) \in \mathbb{R}^{m \times d_y}$ is the output prediction matrix of a neural network, and $\theta \in \mathbb{R}^{d_\theta}$ is the vector consisting of all trainable parameters. Here, $\frac{2}{m} L(\theta)$ is the standard mean squared error, for which all of our results hold true as well, because multiplying $L(\theta)$ by a constant $\frac{2}{m}$ (in $\theta$) changes only the entire scale of the optimization landscape.

The output prediction matrix $\hat{Y}(X, \theta) \in \mathbb{R}^{m \times d_y}$ is specified for shallow networks with rectified linear units (ReLUs) in section 3 and generalized to deep nonlinear neural networks in section 4.

**2.2 Additional Notation.** Define $P[M]$ to be the orthogonal projection matrix onto the column space (or range space) of a matrix $M$. Let $P_N[M]$ be the orthogonal projection matrix onto the null space (or kernel space) of a matrix $M^\top$. For a matrix $M \in \mathbb{R}^{d \times d'}$, we denote the standard vectorization of the matrix $M$ as $\text{vec}(M) = [M_{1,1}, \ldots, M_{d,1}, M_{1,2}, \ldots, M_{d,2}, \ldots, M_{1,d'}, \ldots, M_{d,d'}]^\top$.

## 3 Shallow Nonlinear Neural Networks with Scalar-Valued Output

Before presenting our main results for deep nonlinear neural networks, this section provides the results for shallow networks with a single hidden layer (or three-layer networks with the input and output layers) and scalar-valued output (i.e., $d_y = 1$) to illustrate some of the ideas behind the discussed effects of the depth and width on local minima.

In this section, the vector $\theta \in \mathbb{R}^{d_\theta}$ of all trainable parameters determines the entries of the weight matrices $W^{(1)} := W^{(1)}(\theta) \in \mathbb{R}^{d_x \times d}$ and $W^{(2)} := W^{(2)}(\theta) \in \mathbb{R}^d$ as $\text{vec}([W^{(1)}(\theta), W^{(2)}(\theta)]) = \theta$. Given an input matrix $X$ and a parameter vector $\theta$, the output prediction matrix $\hat{Y}(X, \theta) \in \mathbb{R}^m$ of a fully connected feedforward network with a single hidden layer can be written as

$$\hat{Y}(X, \theta) := \sigma(XW^{(1)})W^{(2)}, \tag{3.1}$$

where $\sigma : \mathbb{R}^{m \times d} \to \mathbb{R}^{m \times d}$ is defined by coordinate-wise nonlinear activation functions $\sigma_{i,j}$ as $(\sigma(M))_{i,j} := \sigma_{i,j}(M_{i,j})$ for each $(i, j)$.

**3.1 Analysis with ReLU Activations.** In this section, the nonlinear activation function $\sigma_{i,j}$ is assumed to be ReLU as $\sigma_{i,j}(z) = \max(0, z)$. Let $\Lambda^{1,k} \in \mathbb{R}^{m \times m}$ represent a diagonal matrix with diagonal elements corresponding to the activation pattern of the $k$th unit at the hidden layer over $m$ different samples as for all $i \in \{1, \dots, m\}$ and all $k \in \{1, \dots, d\}$,

$$
\Lambda_{ii}^{1,k} = \begin{cases} 1 & \text{if } (XW^{(1)})_{i,k} > 0 \\ 0 & \text{otherwise} \end{cases} .
$$

Let $\Phi^{(1)} := \Phi^{(1)}(X, \theta) := \sigma(XW^{(1)}) \in \mathbb{R}^{m \times d}$ be the postactivation output of the hidden layer.

Under this setting, proposition 1 provides an equation that holds at local minima and illustrates the effect of width for shallow ReLU neural networks.

**Proposition 1.** *Every differentiable local minimum $\theta$ of L satisfies that*

$$
L(\theta) = \frac{1}{2} \|Y\|_2^2 - \frac{1}{2} \left\| P\left[N_1^{(2)} \Phi^{(1)}\right] Y \right\|_2^2 - \sum_{k=1}^{d} \underbrace{\frac{1}{2} \left\| P\left[N_k^{(1)} D_k^{(1)}\right] Y \right\|_2^2}_{\substack{\geq 0 \\ \text{further improvement as a network gets wider}}} ,
$$

(3.2)

*where $D_k^{(1)} = W_k^{(2)} \Lambda^{1,k} X$. Here, $N_1^{(1)} := I_m$, $N_k^{(1)} := P_N[\bar{Q}_{k-1}^{(1)}]$ for any $k \in \{2, \dots, d\}$, and $N_1^{(2)} := P_N[\bar{Q}_d^{(1)}]$, where $\bar{Q}_k^{(1)} := [Q_1^{(1)}, \dots, Q_k^{(1)}]$ and $Q_k^{(l)} := N_k^{(1)} D_k^{(1)}$ for any $k \in \{1, \dots, d\}$.*

Proposition 1 is an immediate consequence of our general result (see theorem 1) in the next section (the proof is provided in section A.1). In the rest of this section, we provide a proof sketch of proposition 1.

A geometric intuition behind proposition 1 is that a local minimum is a global minimum within a local region in $\mathbb{R}^{d_\theta}$ (i.e., a neighborhood of the local minimum), the dimension of which increases as a network gets wider (or the number of parameters increases). Thus, a local minimum is a global minimum of a search space with a larger dimension for a wider network. One can also see this geometric intuition in an analysis as follows. If $\theta$ is a differentiable local minimum, then $\theta$ must be a critical point and thus,

$$
\nabla_\theta L(\theta) = \left(\nabla_\theta \hat{Y}(X, \theta)\right) \left(\hat{Y}(X, \theta) - Y\right) = 0.
$$

By rearranging this,

$$\left(\nabla_\theta \hat{Y}(X, \theta)\right) \hat{Y}(X, \theta) = \left(\nabla_\theta \hat{Y}(X, \theta)\right) Y, \tag{3.3}$$

where we can already see the power of strong overparameterization in that if the matrix $\nabla_\theta \hat{Y}(X, \theta) \in \mathbb{R}^{d_\theta \times m}$ is left-invertible, $\hat{Y}(X, \theta) = Y$, and hence every differentiable local minimum is a global minimum. Here, $\nabla_\theta \hat{Y}(X, \theta)$ is a $d_\theta$ by $m$ matrix, so significantly increasing $d_\theta$ (strong overparameterization) can ensure the left invertibility.

Beyond the strong overparameterization, we proceed with the proof sketch of proposition 1 by taking advantage of the special neural network structures in $\hat{Y}(X, \theta)$ and $\nabla_\theta \hat{Y}(X, \theta)$. We first observe that $\hat{Y}(X, \theta) = \Phi^{(1)} W^{(1)}$ and $\hat{Y}(X, \theta) = D^{(1)} \text{vec}(W^{(2)})$, where $D^{(1)} := [D_k^{(1)}]_{k=1}^{d_\theta}$. Moreover, at any differentiable point, we have that $\nabla_{W^{(1)}} \hat{Y}(X, \theta) = (\Phi^{(1)})^\top$ and $\nabla_{\text{vec}(W^{(2)})} \hat{Y}(X, \theta) = (D^{(1)})^\top$. Combining these with equation 3.3 yields

$$\begin{bmatrix} \Phi^{(1)} & D^{(1)} \end{bmatrix}^\top \left( \frac{1}{2} \begin{bmatrix} \Phi^{(1)} & D^{(1)} \end{bmatrix} \begin{bmatrix} W^{(1)} \\ \text{vec}(W^{(2)}) \end{bmatrix} \right) = \begin{bmatrix} \Phi^{(1)} & D^{(1)} \end{bmatrix}^\top Y,$$

where

$$\hat{Y}(X, \theta) = \frac{1}{2} \begin{bmatrix} \Phi^{(1)} & D^{(1)} \end{bmatrix} \begin{bmatrix} W^{(1)} \\ \text{vec}(W^{(2)}) \end{bmatrix}.$$

By solving for the vector $\begin{bmatrix} W^{(1)} & \text{vec}(W^{(2)}) \end{bmatrix}$,

$$\hat{Y}(X, \theta) = P \begin{bmatrix} [ D^{(1)} & \Phi^{(1)} ] \end{bmatrix} Y.$$

Therefore,

$$L(\theta) = \frac{1}{2} \left\| Y - P \begin{bmatrix} [ D^{(1)} & \Phi^{(1)} ] \end{bmatrix} Y \right\|_2^2 = \|Y\|_2^2 - \left\| P \begin{bmatrix} [ D^{(1)} & \Phi^{(1)} ] \end{bmatrix} Y \right\|_2^2,$$

where the second equality follows the idempotence of the projection. Finally, decomposing the second term $\| P \begin{bmatrix} [ D^{(1)} & \Phi^{(1)} ] \end{bmatrix} Y \|_2^2$ by following the Gram-Schmidt process on the set of column vectors of $[ D^{(1)} \; \Phi^{(1)} ]$ yields the desired statement of proposition 1, completing its proof sketch. In proposition 1, the matrices $N_k^{(l)}$ (and $Q_k^{(l)}$) are by-products of this Gram-Schmidt process.

## 3.2 Probabilistic Bound.

From equation 2.2 in proposition 1, the loss $L(\theta)$ at differentiable local minima is expected to tend to get smaller as the width of the hidden layer $d$ gets larger. To further support this theoretical observation, this section obtains a probabilistic upper bound on the loss $L(\theta)$ for white noise data by fixing the activation patterns $\Lambda^{1,k}$ for $k \in \{1, 2, \ldots, d\}$ and assuming that the data matrix $[X \quad Y]$ is a random gaussian matrix, with each entry having mean zero and variance one.

In this section, each nonlinear activation function $\sigma_{i,j}$ is assumed to be ReLU ($\sigma_{i,j}(z) = \max(0, z)$) and leaky ReLU ($\sigma_{i,j}(z) = \max(az, z)$ with any fixed $a \leq 1$) or absolute value activation ($\sigma_{i,j}(z) = |z|$). Let $\Lambda^{1,k} \in \mathbb{R}^{m \times m}$ represent a diagonal matrix with diagonal elements corresponding to the activation pattern of the $k$th unit at the hidden layer over $m$ different samples as

$$\Lambda_{ii}^{1,k} := \begin{cases} \left.\dfrac{\partial \sigma_{i,k}^{(1)}(z)}{\partial z}\right|_{z=(XW^{(1)})_{i,k}} & \text{if } \left.\dfrac{\partial \sigma_{i,k}^{(1)}(z)}{\partial z}\right|_{z=(XW^{(1)})_{i,k}} \text{ exists} \\ 0 & \text{otherwise} \end{cases}.$$

This definition of $\Lambda_{ii}^{1,k}$ generalizes the corresponding definition in section 3.1. Proposition 1 holds for this generalized activation pattern by simply replacing the previous definition of $\Lambda_{ii}^{1,k}$ by this more general definition. This can be seen from the proof sketch in section 3.1 and is later formalized in the proof of theorem 1.

We denote the vector consisting of the diagonal entries of $\Lambda^{1,k}$ by $\Lambda^k \in \mathbb{R}^m$ for $k \in \{1, 2, \ldots, d\}$. Define the activation pattern matrix as $\Lambda := [\Lambda^k]_{k=1}^d \in \mathbb{R}^{m \times d}$. For any index set $I \subseteq \{1, 2, \ldots, m\}$, let $\Lambda_I$ denote the submatrix of $\Lambda$ that consists of its rows of indices in $I$. Let $s_{\min}(\Lambda_I)$ be the smallest singular value of $\Lambda_I$.

Proposition 2 proves that $L(\theta) \approx (1 - d_x d/m)\|Y\|_2^2/2$ in the regime $d_x d \ll m$, and $L(\theta) = 0$ in the regime $d_x d \gg m$, under the corresponding conditions on $\Lambda$; that is, $s_{\min}(\Lambda_I) \geq \delta$ for any index set $I \subseteq \{1, 2, \ldots, m\}$ such that $|I| \geq m/2$ in the regime $d_x d \ll m$, and $|I| \leq d/2$ in the regime $d_x d \gg m$. This supports our theoretical observation that increasing width helps improve the quality of local minima.

**Proposition 2.** *Fix the activation pattern matrix* $\Lambda = [\Lambda^k]_{k=1}^d \in \mathbb{R}^{m \times d}$. *Let* $[X \quad Y]$ *be a random* $m \times (d_x + 1)$ *gaussian matrix, with each entry having mean zero and variance one. Then the loss* $L(\theta)$ *as in equation 3.2 satisfies both of the following statements:*

  i. *If* $m \geq 64 \ln^2(d_x dm/\delta^2) d_x d$ *and* $s_{\min}(\Lambda_I) \geq \delta$ *for any index set* $I \subseteq \{1, 2, \ldots, m\}$ *with* $|I| \geq m/2$, *then*

$$L(\theta) \leq \left(1 + 6\sqrt{\frac{t}{m}}\right) \frac{m - d_x d}{2m}\|Y\|_2^2,$$

*with probability at least* $1 - e^{-m/(64 \ln(d_x dm/\delta^2))} - 2e^{-t}$.

   *ii. If $dd_x \geq 2m \ln^2(md/\delta)$ with $d_x \geq \ln^2(dm)$ and $s_{\min}(\Lambda_I) \geq \delta$ for any index set $I \subseteq \{1, 2, \ldots, m\}$ with $|I| \leq d/2$, then*

$$L(\theta) = 0$$

   *with probability at least $1 - 2e^{-d_x/20}$.*

The proof of proposition 2 is provided in appendix B. In that proof, we first rewrite the loss $L(\theta)$ as the projection of $Y$ onto the null space of an $m \times dd_0$ matrix $\tilde{D}$, with an explicit expression in terms of the activation pattern matrix $\Lambda$ and the data matrix $X$. By our assumption, the data matrix $X$ is a random gaussian matrix. The projection matrix $\tilde{D}$ is also a random matrix. Proposition 2 then boils down to understanding the rank of the projection matrix $\tilde{D}$, and we proceed to show that $\tilde{D}$ has the largest possible rank, $\min\{dd_0, m\}$, with high probability. In fact, we derive quantitative estimates on the smallest singular value of $\tilde{D}$. The main difficulties are that the columns of the matrix $\tilde{D}$ are correlated and variances of different entries vary. Our approach to obtain quantitative estimates on the smallest singular value of $\tilde{D}$ combines the epsilon net argument with an iterative argument.

In the regime $dd_0 \gg m$, results similar to proposition 2ii were obtained under certain diversity assumptions on the entries of the weight matrices in a previous study (Xie, Liang, & Song, 2017). When compared with the previous study (Xie et al., 2017), proposition 2 specifies precise relations between the size $dd_0$ of the neural network and the size $m$ of the data set and also holds true in the regime $dd_0 \ll m$. Moreover, our proof arguments for proposition 2ii are different. Xie et al. (2017), under the assumption that $dd_0 \gg m$, show that $\tilde{D}\tilde{D}^T$ is close to its expectation in the sense of spectral norm. As a consequence, the lower bound of the smallest eigenvalue of $\mathbb{E}[\tilde{D}\tilde{D}^T]$ gives the lower bound for the smallest singular value of $\tilde{D}$.

However, proposition 2 assumes a gaussian data matrix, which may be a substantial limitation. The proof of proposition 2 relies on the concentration properties of gaussian distribution. Whereas a similar proof would be able to extend proposition 2 to a nongaussian distribution with these properties (e.g., distributions with subgaussian tails), it would be challenging to use a similar proof for a general distribution without the properties similar to those.

## 4  Deep Nonlinear Neural Networks

Let $H$ be the number of hidden layers and $d_l$ be the width (or, equivalently, the number of units) of the $l$th hidden layer. To theoretically analyze concrete phenomena, the rest of this letter focuses on fully connected feedforward networks with various depths $H \geq 1$ and widths $d_l \geq 1$, using rectified linear units (ReLUs), leaky ReLUs, and absolute value activations, evaluated with the squared loss function. In the rest of this letter, the (finite) depth $H$ can be arbitrarily large and the (finite) widths $d_l$ can arbitrarily differ among different layers.

**4.1 Model and Notation.** Let $\theta \in \mathbb{R}^{d_\theta}$ be the vector consisting of all trainable parameters, which determines the entries of the weight matrix $W^{(l)} := W^{(l)}(\theta) \in \mathbb{R}^{d_{l-1} \times d_l}$ at every $l$th hidden layer as vec$([W^{(l)}(\theta)]_{l=1}^{H+1}) = \theta$. Here, $d_\theta := \sum_{l=1}^{H+1} d_{l-1} d_l$ is the number of trainable parameters. Given an input matrix $X$ and a parameter vector $\theta$, the output prediction matrix $\hat{Y}(X, \theta) \in \mathbb{R}^{m \times d_{H+1}}$ of a fully connected feedforward network can be written as

$$\hat{Y}(X, \theta) := \Phi^{(H)} W^{(H+1)}, \tag{4.1}$$

where $\Phi^{(l)} := \Phi^{(l)}(X, \theta) \in \mathbb{R}^{m \times d_l}$ is the postactivation output of $l$th hidden layer,

$$\Phi^{(l)}(X, \theta) := \sigma^{(l)}(\Phi^{(l-1)} W^{(l)}),$$

where $\Phi^{(0)}(X, \theta) := X$, $\Phi^{(H+1)}(X, \theta) := \hat{Y}(X, \theta)$, and $\sigma^{(l)} : \mathbb{R}^{m \times d_l} \to \mathbb{R}^{m \times d_l}$ is defined by coordinate-wise nonlinear activation functions $\sigma_{i,j}^{(l)}$ as $(\sigma^{(l)}(M))_{i,j} := \sigma_{i,j}^{(l)}(M_{i,j})$ for each $(l, i, j)$. Each nonlinear activation function $\sigma_{i,j}^{(l)}$ is allowed to differ among different layers and different units within each layer, but assumed to be ReLU ($\sigma_{i,j}^{(l)}(z) = \max(0, z)$), leaky ReLU ($\sigma_{i,j}^{(l)}(z) = \max(az, z)$ with any fixed $a \le 1$) or absolute value activation ($\sigma_{i,j}^{(l)}(z) = |z|$). Here, $d_{H+1} = d_y$ and $d_0 = d_x$. Let $\Lambda^{l,k} \in \mathbb{R}^{m \times m}$ represent a diagonal matrix with diagonal elements corresponding to the activation pattern of the $k$th unit at the $l$th layer over $m$ different samples as

$$\Lambda_{ii}^{l,k} := \begin{cases} \dfrac{\partial \sigma_{i,k}^{(l)}(z)}{\partial z}\bigg|_{z=(\Phi^{(l-1)} W^{(l)})_{i,k}} & \text{if } \dfrac{\partial \sigma_{i,k}^{(l)}(z)}{\partial z}\bigg|_{z=(\Phi^{(l-1)} W^{(l)})_{i,k}} \text{ exists} \\ 0 & \text{otherwise} \end{cases}.$$

This definition of $\Lambda_{ii}^{l,k}$ generalizes the corresponding definition in section 3. Let $I_d$ be the identity matrix of size $d$ by $d$. Define $M \otimes M'$ to be the Kronecker product of matrices $M$ and $M'$. Given a matrix $M$, $M_{\cdot,j}$ and $M_{i,\cdot}$ denote the $j$th column vector of $M$ and the $i$th row vector of $M$, respectively.

**4.2 Theoretical Result.** For the standard deep nonlinear neural networks, theorem 1 provides an equation that holds at local minima and illustrates the effect of depth and width. Let $d'_l := d_l$ for all $l \in \{1, \ldots, H\}$ and $d'_{H+1} := 1$.

**Theorem 1.** *Every differentiable local minimum $\theta$ of $L$ satisfies that*

$$L(\theta) = \frac{1}{2} \|Y\|_F^2 - \sum_{l=1}^{H+1} \sum_{k_l=1}^{d_l'} \underbrace{\frac{1}{2} \left\| P\left[N_{k_l}^{(l)} D_{k_l}^{(l)}\right] vec(Y) \right\|_2^2}_{\substack{\geq 0 \\ \text{\tiny further improvement as} \\ \text{\tiny a network gets wider and deeper}}}, \tag{4.2}$$

*where $D_{k_l}^{(l)} := D_{k_l}^{(l)}(\theta)$ and $N_{k_l}^{(l)} := N_{k_l}^{(l)}(\theta)$ are defined as follows. For any $l \in \{1, \ldots, H\}$ and any $k_l \in \{1, \ldots, d_l\}$,*

$$D_{k_l}^{(l)} := \sum_{k_{l+1}=1}^{d_{l+1}} \cdots \sum_{k_H=1}^{d_H} (W_{k_l,k_{l+1}}^{(l+1)} \cdots W_{k_{H-1},k_H}^{(H)} W_{k_H,\cdot}^{(H+1)})^\top \otimes \Lambda^{l,k_l} \cdots \Lambda^{H,k_H} \Phi^{(l-1)},$$

*with $D_{k_H}^{(H)} := (W_{k_H,\cdot}^{(H+1)})^\top \otimes \Lambda^{H,k_H} \Phi^{(H-1)}$. For any $l \in \{1, \ldots, H\}$ and any $k_l \in \{1, \ldots, d_l\}$, $N_{k_l}^{(l)} := P_N[\bar{Q}_{k_l-1}^{(l)}]$ with $N_1^{(1)} := I_m$ where $\bar{Q}_{k_l}^{(l)} := [Q_1^{(1)}, \ldots, Q_{d_1}^{(1)}, Q_1^{(2)}, \ldots, Q_{d_2}^{(2)}, \ldots, Q_1^{(l)}, \ldots, Q_{k_l}^{(l)}]$, $Q_{k_l}^{(l)} := N_{k_l}^{(l)} D_{k_l}^{(l)}$, and $\bar{Q}_0^{(l)} := \bar{Q}_{d_{l-1}}^{(l-1)}$. Here, $D_1^{(H+1)}(\theta) := I_{d_{H+1}} \otimes \Phi^{(H)}$ and $N_1^{(H+1)}(\theta) := P_N[\bar{Q}_{d_H}^{(H)}]$.*

The complete proof of theorem 1 is provided in section A.1. Theorem 1 is a generalization of proposition 1. Accordingly, its proof follows the proof sketch presented in the previous section for proposition 1.

Unlike previous studies (Livni et al., 2014; Nguyen & Hein, 2017, 2018), theorem 1 requires no overparameterization such as $d_l \geq m$. Instead, it provides quantitative gradual effects of depth and width on local minima, from no overparameterization to overparameterization. Notably, theorem 1 shows the effect of overparameterization in terms of depth as well as width, which also differs from the results of previous studies that consider overparameterization in terms of width (Livni et al., 2014; Nguyen & Hein, 2017, 2018).

The proof idea behind these previous studies with strong overparameterization is captured in the discussion after equation 3.3—with strong overparameterization such that $d_l \geq m$ and $\text{rank}(D^{(1)}) \geq m$, $\nabla_{\text{vec}(W)} \hat{Y}(X, \theta) \in \mathbb{R}^{d_l \times m}$ is left-invertible and hence every local minimum is a global minimum with zero training error. Here, $\text{rank}(M)$ represents the rank of a matrix $M$. The proof idea behind theorem 1 differs from those as shown in section 3.1. What is still missing in theorem 1 is the ability to provide a prior guarantee on $L(\theta)$ without strong overparameterization, which is addressed in sections 3.2 and 5 for some special cases but left as an open problem for other cases.

**4.3 Experiments.** In theorem 1, we have shown that at every differentiable local minimum $\theta$, the total training loss value $L(\theta)$ has an analytical formula $L(\theta) = J(\theta)$, where

$$J(\theta) := \frac{1}{2} \|Y\|_F^2 - \sum_{l=1}^{H+1} \sum_{k_l=1}^{d_l'} \frac{1}{2} \left\| P \left[ N_{k_l}^{(l)}(\theta) D_{k_l}^{(l)}(\theta) \right] \text{vec}(Y) \right\|_2^2$$

denotes the right-hand side of equation 4.1. In this section, we investigate the actual numerical values of the formula $J(\theta)$ with a synthetic data set and standard benchmark data sets for neural networks with different degrees of depth $= H$ and hidden layers' width $= d_l$ for $l \in \{1, 2, \ldots, H\}$.

In the synthetic data set, the data points $\{(x_i, y_i)\}_{i=1}^m$ were randomly generated by a ground-truth, fully connected feedforward neural network with $H = 7, d_l = 50$ for all $l \in \{1, 2, \ldots, H\}$, tanh activation function, $(x, y) \in \mathbb{R}^{10} \times \mathbb{R}$ and $m = 5000$. MNIST (LeCun, Bottou, Bengio, & Haffner, 1998), a popular data set for recognizing handwritten digits, contains $28 \times 28$ grayscale images. The CIFAR-10 (Krizhevsky & Hinton, 2009) data set consists of $32 \times 32$ color images that contain different types of objects such as "airplane," "automobile," and "cat." The Street View House Numbers (SVHN) data set (Netzer et al., 2011) contains house digits collected by Google Street View, and we used the $32 \times 32$ color image version for the standard task of predicting the digits in the middle of these images. In order to reduce the computational cost, for the image data sets (MNIST, CIFAR-10, and SVHN), we center-cropped the images ($24 \times 24$ for MNIST and $28 \times 28$ for CIFAR-10 and SVHN), then resized them to smaller gray-scale images ($8 \times 8$ for MNIST and $12 \times 12$ for CIFAR-10 and SVHN), and used randomly selected subsets of the data sets with size $m = 10,000$ as the training data sets.

For all the data sets, the network architecture was fixed to be a fully connected feedforward network with the ReLU activation function. For each data set, the values of $J(\theta)$ were computed with initial random weights drawn from a normal distribution with zero mean and normalized standard deviation $(1/\sqrt{d_l})$ and with trained weights at the end of 40 training epochs. (Additional experimental details are presented in appendix C.)

Figure 1 shows the results with the synthetic data set, as well as the MNIST, CIFAR-10, and SVHN data sets. As it can be seen, the values of $J(\theta)$ tend to decrease toward zero (and hence the global minimum value), as the width or depth of neural networks increases. In theory, the values of $J(\theta)$ may not improve as much as desired along depth and width if representations corresponding to each unit and each layer are redundant in the sense of linear dependence of the columns of $D_{k_l}^{(l)}(\theta)$ (see theorem 1). Intuitively, at initial random weights, one can mitigate this redundancy due to the randomness of the weights, and hence a major concern is whether such redundancy arises and $J(\theta)$ degrades along with training. From Figure 1, it can be also noticed that the values of $J(\theta)$ tend to decrease along with training. These empirical results partially support our theoretical observation that increasing the depth and width can improve the quality of local minima.

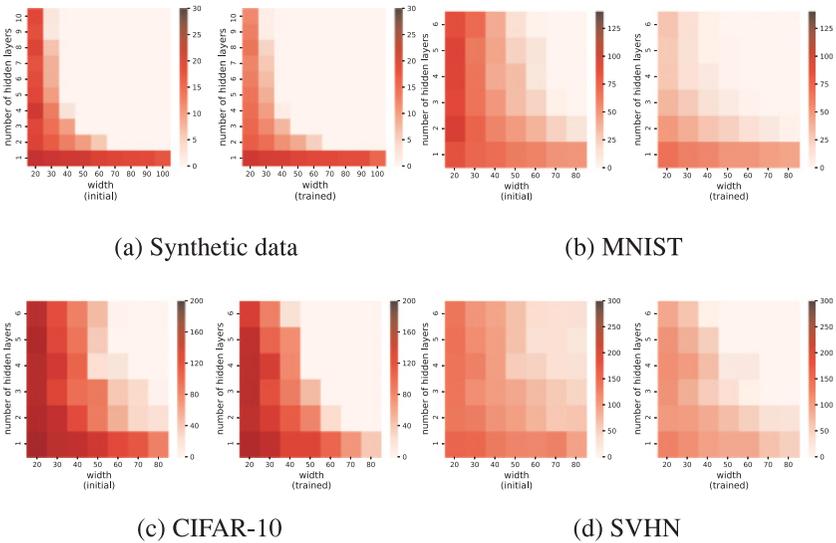

(a) Synthetic data                                          (b) MNIST

(c) CIFAR-10                                                (d) SVHN

Figure 1: The values of $\sqrt{J(\theta)}$ for the training data sets ($J(\theta)$ are on the right-hand side of equation 4.1) with varying depth = $H$ ($y$-axis) and width = $d_l$ for all $l \in \{1, 2, \ldots, H\}$ ($x$-axis). The heat map colors represent the values of $\sqrt{J(\theta)}$. In all panels of this figure, the left heat map (initial) is computed with initial random weights and the right heat map (trained) is calculated after training. It can be seen that both depth and width helped improve the values of $J(\theta)$.

## 5 Deep Nonlinear Neural Networks with Local Structure

Given the scarcity of theoretical understanding of the optimality of deep neural networks, Goodfellow, Bengio, and Courville (2016) noted that it is valuable to theoretically study simplified models: deep linear neural networks. For example, Saxe, McClelland, and Ganguli (2014) empirically showed that in terms of optimization, deep linear networks exhibited several properties similar to those of deep nonlinear networks. Following these observations, the theoretical study of deep linear neural networks has become an active area of research (Kawaguchi, 2016; Hardt & Ma, 2017; Arora, Cohen, Golowich, & Hu, 2018; Arora, Cohen, & Hazan, 2018), as a step toward the goal of establishing the optimization theory of deep learning.

As another step toward the goal, this section discards the strong linearity assumption and considers a locally induced nonlinear-linear structure in deep nonlinear networks with the piecewise linear activation functions such as ReLUs, leaky ReLUs, and absolute value activations.

**5.1 Locally Induced Nonlinear-Linear Structure.** In this section, we describe how a standard deep nonlinear neural network can induce nonlinear-linear structure. The nonlinear-linear structure considered in this letter is defined in definition 1: condition i simply defines the index subsets $S^{(l)}$ that pick out the relevant subset of units at each layer $l$, condition ii requires the existence of $n$ linearly acting units, and condition iii imposes weak separability of edges.

**Definition 1.** *A parameter vector $\theta$ is said to induce $(n, t)$ weakly separated linear units on a training input data set $X$ if there exist $(H + 1 - t)$ sets $S^{(t+1)}, S^{(t+2)}, \ldots, S^{(H+1)}$ such that for all $l \in \{t + 1, t + 2, \ldots, H + 1\}$, the following three conditions hold:*

  i. *$S^{(l)} \subseteq \{1, \ldots, d_l\}$ with $|S^{(l)}| \geq n$.*
  ii. *$\Phi^{(l)}(X, \theta)._{,k} = \Phi^{(l-1)}(X, \theta)W^{(l)}(\theta)._{,k}$ for all $k \in S^{(l)}$.*
  iii. *$W^{(l+1)}(\theta)_{k',k} = 0$ for all $(k', k) \in S^{(l)} \times (\{1, \ldots, d_{l+1}\} \setminus S^{(l+1)})$ if $l \leq H - 1$.*

Given a training input data set $X$, let $\Theta_{n,t}$ be the set of all parameter vectors that induce $(n, t)$ weakly separated linear units on the training input data set $X$ that defines the total loss $L(\theta)$ in equation 2.1. For standard deep nonlinear neural networks, all parameter vectors $\theta$ are in $\Theta_{d_{H+1},H}$, and some parameter vectors $\theta$ are in $\Theta_{n,t}$ for different values of $(n, t)$. Figure 2 a illustrates locally induced structures for $\theta \in \Theta_{1,0}$. For a parameter $\theta$ to be in $\Theta_{n,t}$, definition 1 requires only the existence of a portion $n/d_l$ of units to act linearly on the particular training data set merely at the particular $\theta$. Thus, all units can be nonlinear, act nonlinearly on the training data set outside of some parameters $\theta$, and operate nonlinearly always on other inputs $x$—for example, in a test data set or a different training data set. The weak separability requires that the edges going from the $n$ units to the rest of the network are negligible. The weak separability does not require the $n$ units to be separated from the rest of the neural network.

Here, a neural network with $\theta \in \Theta_{n,t}$ can be a standard deep nonlinear neural network (without any linear units in its architecture), a deep linear neural network (with all activation functions being linear), or a combination of these cases. Whereas a standard deep nonlinear neural network can naturally have parameters $\theta \in \Theta_{n,t}$, it is possible to guarantee all parameters $\theta$ to be in $\Theta_{n,t}$ with desired $(n, t)$ simply by using corresponding network architectures. For standard deep nonlinear neural networks, one can also restrict all relevant convergent solution parameters $\theta$ to be in $\Theta_{n,t}$ by using

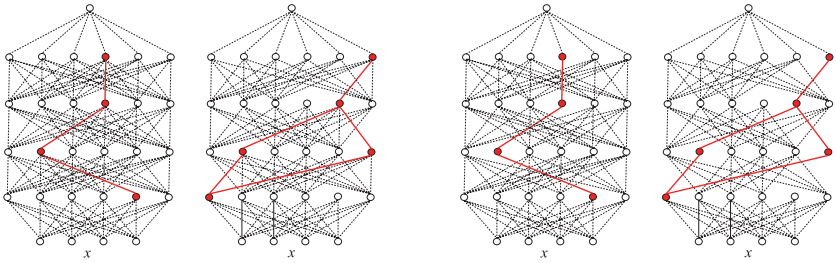

(a) weakly-separated: $\theta \in \Theta_{1,0}$　　(b) strongly-separated: $\theta \in \Theta_{1,0}^{\text{strong}}$

Figure 2: Illustration of locally induced nonlinear-linear structures. (a) Simple examples of the structure with weakly separated edges considered in this section (see definition 1). (b) Examples of a simpler structure with strongly separated edges (see definition 2). The red nodes represent the linearly acting units on a training data set at a particular $\theta$, and the white nodes are the remaining units. The black dashed edges represent standard edges without any assumptions. The red nodes are allowed to depend on all nodes from the previous layer in panel a, whereas they are not allowed in panel b except for the input layer. In both panels a and b, two examples of parameters $\theta$ are presented with the exact same network architecture (including activation functions and edges). Even if the network architecture (or parameterization) is identical, different parameters $\theta$ can induce different local structures. With $\Theta_{1,4}$, this local structure always holds in standard deep nonlinear networks with four hidden layers.

some corresponding learning algorithms. Our theoretical results hold for all of these cases.

**5.2 Theoretical Result.** We state our main theoretical result in theorem 2 and corollary 1; a simplified statement is presented in remark 1. Here, a classical machine learning method, basis function regression, is used as a baseline to be compared with neural networks. The global minimum value of basis function regression with an arbitrary basis matrix $M(X)$ is $\inf_R \frac{1}{2}\|M(X)R - Y\|_F^2$, where the basis matrix $M(X)$ does not depend on $R$ and can represent nonlinear maps, for example, by setting $M = ([\phi(x_i)]_{i=1}^m)^\top \in \mathbb{R}^{m \times d_\phi}$ with any nonlinear basis functions $\phi$ and any finite $d_\phi$. In theorem 2, the expression $P_N\left[\Phi^{(S)}\right]Y$ represents the projection of $Y$ onto the null space of $(\Phi^{(S)})^\top$, which is also ($Y$—the projection of $Y$ onto the column space of $\Phi^{(S)}$). Given matrices $(M^{(j)})_{j \in S}$ with a sequence $S = (s_1, s_2, \ldots, s_n)$, define $[M^{(j)}]_{j \in S} := \begin{bmatrix} M^{(s_1)} & M^{(s_2)} & \cdots & M^{(s_n)} \end{bmatrix}$ to be a block matrix with columns being $M^{(s_1)}, M^{(s_2)}, \ldots, M^{(s_n)}$. Let $S \subseteq (s_1, s_2, \ldots, s_n)$ denote a subsequence of $(s_1, s_2, \ldots, s_n)$.

**Theorem 2.** *For any $t \in \{0, 1, \ldots, H\}$, every differentiable local minimum $\theta \in \Theta_{d_{H+1}, t}$ of $L$ satisfies that for any subsequence $S \subseteq (t, t+1, \ldots, H)$ (including the case of $S$ being the empty sequence),*

$$L(\theta) \leq \underbrace{\frac{1}{2} \left\| P_N \left[ \Phi^{(S)} \right] Y \right\|_F^2}_{\substack{\text{global minimum value of} \\ \text{basis function regression} \\ \text{with basis matrix } \Phi^{(S)}}} - \underbrace{\sum_{l=1}^{H} \sum_{k_l=1}^{d_l} \underbrace{\frac{1}{2} \left\| P \left[ N_{k_l}^{(l)} P_N \left[ \bar{\Phi}^{(S)} \right] D_{k_l}^{(l)} \right] vec(Y) \right\|_2^2}_{\geq 0}}_{\substack{\text{further improvement as} \\ \text{a network gets wider and deeper}}},$$

(5.1)

*where $P_N \left[ \Phi^{(S)} \right] \in \mathbb{R}^{m \times m}$, $P_N \left[ \bar{\Phi}^{(S)} \right] \in \mathbb{R}^{m d_{H+1} \times m d_{H+1}}$, $\Phi^{(S)} = [\Phi^{(l)}]_{l \in S}$, $\bar{\Phi}^{(S)} = [I_{d_{H+1}} \otimes \Phi^{(l)}]_{l \in S}$. If $S$ is empty, $P_N[\Phi^{(S)}] = I_m$ and $P_N[\bar{\Phi}^{(S)}] = I_{m d_{H+1}}$. The matrices $D_{k_l}^{(l)}$ and $N_{k_l}^{(l)}$ are defined in theorem 1 with the exception that $Q_{k_l}^{(l)} = N_{k_l}^{(l)} P_N[\bar{\Phi}^{(S)}] D_{k_l}^{(l)}$ (instead of $Q_{k_l}^{(l)} := N_{k_l}^{(l)} D_{k_l}^{(l)}$).*

**Remark 1.** From theorem 2 (or corollary 1), one can see the following properties of the loss landscape:

  i. Every differentiable local minimum, $\theta \in \Theta_{d_{H+1}, t}$ has a loss value $L(\theta)$ better than or equal to any global minimum value of basis function regression with any combination of the basis matrices in the set $\{\Phi^{(l)}\}_{l=t}^{H}$ of fixed deep hierarchical representation matrices. In particular with $t = 0$, every differentiable local minimum $\theta \in \Theta_{d_{H+1}, 0}$ has a loss value $L(\theta)$ no worse than the global minimum values of standard basis function regression with the handcrafted basis matrix $\Phi^{(0)} = X$, and of basis function regression with the larger basis matrix $[\Phi^{(l)}]_{l=0}^{H}$.

  ii. As $d_l$ and $H$ increase (or, equivalently, as a neural network gets wider and deeper), the upper bound on the loss values of local minima can further improve.

The proof of theorem 2 is provided in section A.2. The proof is based on the combination of the idea presented in section 3.1 and perturbations of a local minimum candidate. That is, if a $\theta$ is a local minimum, then the $\theta$ is a global minimum within a local region (i.e., a neighborhood of $\theta$). Thus, after perturbing $\theta$ as $\theta' = \theta + \Delta\theta$ such that $\|\Delta\theta\|$ is sufficiently small (so that $\theta'$ stays in the local region) and $L(\theta') = L(\theta)$, the $\theta'$ must be still a global minimum within the local region and, hence, the $\theta'$ is also a local minimum. The proof idea of theorem 2 is to apply the proof sketch in section 3.1 to not only a local minimum candidate $\theta$ but also its perturbations $\theta' = \theta + \Delta\theta$.

In terms of overparameterization, theorem 2 states that local minima of deep neural networks are as good as global minima of the corresponding basis function regression even without overparameterization, and overparameterization helps to further improve the guarantee on local minima. The effect of overparameterization is captured in both the first and second terms on the right-hand side of equation 5.1. As depth and width increase, the second term tends to increase, and hence the guarantee on local minima can improve. Moreover, as depth and width increase (for some of $t + 1, t + 2, \ldots, H$th layers in theorem 2), the first term tends to decrease and the guarantee on local minima can also improve. For example, if $[\Phi^{(l)}]_{l=t}^H$ has rank at least $m$, then the first term is zero and, hence, every local minimum is a global minimum with zero loss value. As a special case of this example, since every $\theta$ is automatically in $\Theta_{d_{H+1}, H}$, if $\Phi^{(H)}$ is forced to have rank at least $m$, every local minimum becomes a global minimum for standard deep nonlinear neural networks, which coincides with the observation about overparameterization by Livni et al. (2014).

Without overparameterization, theorem 2 also recovers one of the main results in the literature of deep linear neural networks as a special case—that is, every local minimum is a global minimum. If $d_{H+1} \leq \min\{d_l : 1 \leq l \leq H\}$, every local minimum $\theta$ for deep linear networks is differentiable and in $\Theta_{d_{H+1}, 0}$, and hence theorem 1 yields that $L(\theta) \leq \frac{1}{2}\|P_N[X]Y\|_F^2$. Because $\frac{1}{2}\|P_N[X]Y\|_F^2$ is the global minimum value, this implies that every local minimum is a global minimum for deep linear neural networks.

Corollary 1 states that the same conclusion and discussions as in theorem 2 hold true even if we fix the edges in condition iii in definition 1 to be zero (by removing them as an architectural design or by forcing it with a learning algorithm) and consider optimization problems only with remaining edges.

**Corollary 1.** *For any $t \in \{0, 1, \ldots, H\}$, every differentiable local minimum $\theta \in \Theta_{d_{H+1}, t}$ of $L|_{\mathcal{I}}$ satisfies that for any subsequence $S \subseteq (t, t + 1, \ldots, H)$ (including the case of $S$ being the empty sequence),*

$$L(\theta) \leq \underbrace{\frac{1}{2}\left\|P_N\left[\Phi^{(S)}\right]Y\right\|_F^2}_{\substack{\text{global minimum value of} \\ \text{basis function regression} \\ \text{with basis matrix } \Phi^{(S)}}} - \underbrace{\sum_{l=1}^{H}\sum_{k_l=1}^{d_l}\underbrace{\frac{1}{2}\left\|P\left[\hat{N}_{k_l}^{(l)}P_N\left[\bar{\Phi}^{(S)}\right]\hat{D}_{k_l}^{(l)}\right]vec(Y)\right\|_2^2}_{\geq 0}}_{\substack{\text{further improvement as} \\ \text{a network gets wider and deeper}}},$$

(5.2)

*where $L|_{\mathcal{I}}$ is the restriction of $L$ to $\mathcal{I} = \{\theta' \in \mathbb{R}^{d_\theta} : \forall l \in \{t + 1, \ldots, H - 1\}, \forall(k', k) \in S^{(l)} \times \overline{S^{(l+1)}}, W^{(l+1)}(\theta')_{k',k} = 0\}$ with the index sets $S^{(t+1)}, S^{(t+2)}, \ldots, S^{(H+1)}$ of the $\theta \in \Theta_{d_{H+1}, t}$ in definition 1 and $\overline{S^{(l)}} := \{1, \ldots, d_l\} \setminus S^{(l)}$. Here, $\Phi^{(S)}$ and $\bar{\Phi}^{(S)}$ are defined in theorem 2, and the matrices $\hat{D}_{k_l}^{(l)}$ and $\hat{N}_{k_l}^{(l)}$ are defined as follows. For all $l \in \{1, \ldots, t + 1\}, \hat{D}_{k_l}^{(l)} := D_{k_l}^{(l)}$ for all $k_l \in \{1, \ldots, d_l\}$ (where $D_{k_l}^{(l)}$ is defined in theorem 2). For all $l \in \{t + 2, \ldots, H\}, \hat{D}_{k_l}^{(l)} := D_{k_l}^{(l)}$ for all $k_l \in S^{(l)}$, and*

$$\hat{D}_{k_l}^{(l)} := \sum_{k_{l+1}=1}^{d_{l+1}} \cdots \sum_{k_H=1}^{d_H} (W_{k_l,k_{l+1}}^{(l+1)} \cdots W_{k_{H-1},k_H}^{(H)} W_{k_H,\cdot}^{(H+1)})^\top$$

$$\otimes \Lambda^{l,k_l} \cdots \Lambda^{H,k_H} [\Phi_{\cdot,j}^{(l-1)}]_{j \in \overline{S^{(l-1)}}},$$

with $\hat{D}_{k_l}^{(H)} := (W_{k_H,\cdot}^{(H+1)})^\top \otimes \Lambda^{H,k_H} [\Phi_{\cdot,j}^{(H-1)}]_{j \in \overline{S^{(H-1)}}}$ for all $k_l \in \overline{S^{(l)}}$. For any $l \in \{1, \ldots, H\}$ and any $k_l \in \{1, \ldots, d_l\}$, $\hat{N}_{k_l}^{(l)} := P_N[\bar{Q}_{k_l-1}^{(l)}]$ with $N_1^1 := I_m$ where $\bar{Q}_{k_l}^{(l)} := [Q_1^{(1)}, \ldots, Q_{d_1}^{(1)}, Q_1^{(2)}, \ldots, Q_{d_2}^{(2)}, \ldots, Q_1^{(l)}, \ldots, Q_{k_l}^{(l)}]$, $Q_{k_l}^{(l)} := \hat{N}_{k_l}^{(l)} P_N[\bar{\Phi}^{(S)}] \hat{D}_{k_l}^{(l)}$, and $\bar{Q}_0^{(l)} := \bar{Q}_{d_{l-1}}^{(l-1)}$.

The proof of corollary 1 is provided in section A.3 and follows the proof of theorem 2. Here, $\Phi^{(0)} = X$ consists of training inputs $x_i$ in the arbitrary given feature space embedded in $\mathbb{R}^{d_0}$; for example, given a raw input $x^{\text{raw}}$ and any feature map $\phi : x^{\text{raw}} \mapsto \phi(x^{\text{raw}}) \in \mathbb{R}^{d_0}$ (including identity as $\phi(x^{\text{raw}}) = x^{\text{raw}}$), we write $x = \phi(x^{\text{raw}})$. Therefore, theorem 2 and corollary 1 state that every differentiable local minima of deep neural networks can be guaranteed to be no worse than any given basis function regression model with a handcrafted basis taking values in $\mathbb{R}^d$ with some finite $d$, such as polynomial regression with a finite degree and radial basis function regression with a finite number of centers.

To illustrate an advantage of the notion of weakly separated edges in definition 1, one can consider the following alternative definition that requires strongly separated edges.

**Definition 2.** *A parameter vector $\theta$ is said to induce $(n, t)$ strongly separated linear units on the training input data set $X$ if there exist $(H + 1 - t)$ sets $S^{(t+1)}, S^{(t+2)}, \ldots, S^{(H+1)}$ such that for all $l \in \{t + 1, t + 2, \ldots, H + 1\}$, conditions i to iii in definition 1 hold and $\Phi^{(l)}(X, \theta) W^{(l+1)}(\theta)_{\cdot,k} = \sum_{k' \in S^{(l)}} \Phi^{(l)}(X, \theta)_{\cdot,k'} W^{(l+1)}(\theta)_{k',k}$ for all $k \in S^{(l+1)}$ if $l \neq \{H, H+1\}$.*

Let $\Theta_{n,t}^{\text{strong}}$ be the set of all parameter vectors that induces $(n, t)$ strongly-separated linear units on the particular training input data set $X$ that defines the total loss $L(\theta)$ in equation 2.1. Figure 2 shows a comparison of weakly separated edges and strongly separated edges. Under this stronger restriction on the local structure, we can obtain corollary 2.

**Corollary 2.** *For any $t \in \{0, 1, \ldots, H\}$, every differentiable local minimum $\theta \in \Theta_{d_{H+1},t}^{\text{strong}}$ of $L$ satisfies that for any $S \subseteq (t, H)$,*

$$L(\theta) \leq \frac{1}{2} \left\| P_N \left[ \Phi^{(S)} \right] Y \right\|_F^2 - \sum_{l=1}^{H} \sum_{k_l=1}^{d_l} \frac{1}{2} \left\| P \left[ N_{k_l}^{(l)} P_N \left[ \bar{\Phi}^{(S)} \right] D_{k_l}^{(l)} \right] vec(Y) \right\|_2^2,$$

*where $\Phi^{(S)}$, $\bar{\Phi}^{(S)}$, $D_{k_l}^{(l)}$, and $N_{k_l}^{(l)}$ are defined in theorem 2.*

The proof of corollary 2 is provided in section A.4 and follows the proof of theorem 2. As a special case, corollary 2 also recovers the statement that every local minimum is a global minimum for deep linear neural networks in the same way as in theorem 2. When compared with theorem 2, one can see that the statement in corollary 2 is weaker, producing the upper bound only in terms of $S \subseteq (t, H)$. This is because the restriction of strongly separated units forces neural networks to have less expressive power with fewer effective edges. This illustrates an advantage of the notion of weakly separated edges in definition 1.

A limitation in theorems 1 and 2 and corollary 1 is the lack of treatment of nondifferentiable local minima. The Lebesgue measure of nondifferentiable points is zero, but this does not imply that the appropriate measure of nondifferentiable points is small. For example, if $L(\theta) = |\theta|$, the Lebesgue measure of the nondifferentiable point ($\theta = 0$) is zero, but the nondifferentiable point is the only local and global minimum. Thus, the treatment of nondifferentiable points in this context is a nonnegligible problem. The proofs of theorems 1 and 2 and corollary 1 are all based on the proof sketch in section 3.1, which heavily relies on the differentiability. Thus, the current proofs do not trivially extend to address this open problem.

## 6 Conclusion

In this letter, we have theoretically and empirically analyzed the effect of depth and width on the loss values of local minima, with and without a possible local nonlinear-linear structure. The local nonlinear-linear structure we have considered might naturally arise during training and also is guaranteed to emerge by using specific learning algorithms or architecture designs. With the local nonlinear-linear structure, we have proved that the values of local minima of neural networks are no worse than the global minimum values of corresponding basis function regression and can improve as depth and width increase. In the general case without the possible local structure, we have theoretically shown that increasing the depth and width can improve the quality of local minima, and we empirically supported this theoretical observation. Furthermore, without the local structure but with a shallow neural network and a gaussian data matrix, we have proven the probabilistic bounds on the rates of the improvements on the local minimum values with respect to width. Moreover, we have discussed a major limitation of this letter: all of its the results focus on the differentiable points on the loss surfaces. Additional treatments of the nondifferentiable points are left to future research.

Our results suggest that the values of local minima are not arbitrarily poor (unless one crafts a pathological worst-case example) and can be guaranteed to some desired degree in practice, depending on the degree of overparameterization, as well as the local or global structural assumption. Indeed, a structural assumption, namely the existence of an identity map, was recently used to analyze the quality of local minima (Shamir, 2018; Kawaguchi & Bengio, 2018). When compared with these previous studies (Shamir, 2018; Kawaguchi & Bengio, 2018), we have shown the effect of depth and width, as well as considered a different type of neural network without the explicit identity map.

In practice, we often "overparameterize" a hypothesis space in deep learning in a certain sense (e.g., in terms of expressive power). Theoretically, with strong overparameterization assumptions, we can show that every stationary point (including all local minima) with respect to a single layer is a global minimum with the zero training error and can memorize any data set. However, "overparameterization" in practice may not satisfy such strong overparameterization assumptions in the theoretical literature. In contrast, our results in this letter do not require overparameterization and show the gradual effects of overparameterization as consequences of general results.

**Appendix A: Proofs for Nonprobabilistic Statements** ───────

Let $D_{k_l}^{(l)}$ be defined in theorem 2. Let $D^{(l)} := [D_k^{(l)}]_{k=1}^{d_l} \in \mathbb{R}^{m d_{H+1} \times d_l d_{l-1}}$ and $D := [D^{(l)}]_{l=1}^{H} \in \mathbb{R}^{m d_{H+1} \times \sum_{l=1}^{H} d_{l-1} d_l}$. Given a matrix-valued function $f(\theta) \in \mathbb{R}^{d' \times d}$, let $\partial_{W^{(l)}} f(\theta) := \frac{\partial \mathrm{vec}(f)}{\partial \mathrm{vec}(W^{(l)})} \in \mathbb{R}^{d' d \times d_{l-1} d_l}$ be the partial derivative of $\mathrm{vec}(f)$ with respect to $\mathrm{vec}(W^{(l)})$. Let $\{j, j+1, \ldots, j'\} := \emptyset$ if $j > j'$. Let $M^{(l)} M^{(l+1)} \cdots M^{(l')} = I$ if $l > l'$. Let $\mathrm{Null}(M)$ be the null space of a matrix $M$. Let $B(\theta, \epsilon)$ be an open ball of radius $\epsilon$ with the center at $\theta$.

The following lemma decomposes the model output $\hat{Y}$ in terms of the weight matrix $W^{(l)}$ and $D^{(l)}$ that coincides with its derivatives at differentiable points.

**Lemma 1.** *For all $l \in \{1, \ldots, H\}$,*

$$vec(\hat{Y}(X, \theta)) = D^{(l)} vec(W^{(l)}(\theta)),$$

*and at any differentiable $\theta$,*

$$\partial_{W^{(l)}} \hat{Y}(X, \theta) = D^{(l)}.$$

**Proof.** Define $G^{(l)}$ to be the preactivation output of the $l$th hidden layer as $G^{(l)} := G^{(l)}(X, \theta) := \Phi^{(l-1)}(X, \theta) W^{(l)}$. By the linearity of the vec operation and the definition of $G^{(l)}$, we have that

$$\mathrm{vec}[G^{(l+1)}(X, \theta)] = \mathrm{vec}\left( \sum_{k=1}^{d_l} \Lambda^{l,k} G^{(l)}(X, \theta)_{\cdot, k} W_{k, \cdot}^{(l+1)} \right)$$

$$= \sum_{k=1}^{d_l} \left( (W_{k,\cdot}^{(l+1)})^\top \otimes \Lambda^{l,k} \right) \text{vec} \left( G^{(l)}(X, \theta)_{\cdot,k} \right)$$

$$= F^{(l+1)} \text{vec} \left( G^{(l)}(X, \theta) \right),$$

where $F^{(l+1)} := [(W_{k,\cdot}^{(l+1)})^\top \otimes \Lambda^{l,k}]_{k=1}^{d_l}$. Therefore,

$$\text{vec}(\hat{Y}) = F^{(H+1)} F^{(H)} \cdots F^{(l+1)} \text{vec}(G^{(l)})$$

$$= F^{(H+1)} \cdots F^{(l+1)} [I_{d_l} \otimes \Phi^{(l-1)}] \text{vec}(W^{(l)}),$$

where $F^{(H+1)} \cdots F^{(l+1)} [I_{d_l} \otimes \Phi^{(l-1)}] = [D_1^{(l)} \quad D_2^{(l)} \quad \cdots \quad D_{d_l}^{(l)}] = D^{(l)}$, which proves the first statement that $\text{vec}(\hat{Y}) = D^{(l)} \text{vec}(W^{(l)})$. The second statement follows from the fact that the derivatives of $D^{(l)}$ with respect to $\text{vec}(W^{(l)})$ are zeros at any differentiable point, and hence $(\partial_{W^{(l)}} \hat{Y}) = D^{(l)} + 0$. □

Lemma 2 generalizes part of theorem A.45 in Rao, Toutenburg, Shalabh, and Heumann (2007) by discarding invertibility assumptions.

**Lemma 2.** *For any block matrix* $[A \ B]$ *with real submatrices A and B such that* $A^\top B = 0$,

$$P[[A \ B]] = P[A] + P[B].$$

**Proof.** It follows a straightforward calculation as

$$P[[A \ B]] = [A \ B] \begin{bmatrix} A^\top A & 0 \\ 0 & B^\top B \end{bmatrix}^\dagger [A \ B]^\top$$

$$= [A \ B] \begin{bmatrix} (A^\top A)^\dagger & 0 \\ 0 & (B^\top B)^\dagger \end{bmatrix} [A \ B]^\top$$

$$= P[A] + P[B].$$

□

Lemma 3 decomposes a norm of a projected target vector into a form that clearly shows an effect of depth and width.

**Lemma 3.** *For any* $t \in \{0, 1, \ldots, H\}$ *and any* $S \subseteq (t, t+1, \ldots, H)$,

$$\left\| P\left[ P_N \left[ \bar{\Phi}^{(S)} \right] D \right] vec(Y) \right\|_2^2 = \sum_{l=1}^{H} \sum_{k_l=1}^{d_l} \left\| P\left[ N_{k_l}^{(l)} P_N \left[ \bar{\Phi}^{(S)} \right] D_{k_l}^{(l)} \right] vec(Y) \right\|_2^2.$$

**Proof.** Since the span of the columns of $[A \; B]$ is the same as the span of the columns of $[A \; P_N[A]B]$ for submatrices $A$ and $B$, the span of the columns of $P_N[\bar{\Phi}^{(S)}]D = [[P_N[\bar{\Phi}^{(S)}]D_{k_l}^{(l)}]_{k_l=1}^{d_l}]_{l=1}^{H}$ is the same as the span of the columns of $[[N_{k_l}^{(l)}P_N[\bar{\Phi}^{(S)}]D_{k_l}^{(l)}]_{k_l=1}^{d_l}]_{l=1}^{H}$. Then, by repeatedly applying lemma 2 to each block of $[[N_{k_l}^{(l)}P_N[\bar{\Phi}^{(S)}]D_{k_l}^{(l)}]_{k_l=1}^{d_l}]_{l=1}^{H}$, we have that

$$P\left[P_N[\bar{\Phi}^{(S)}]D\right] = P\left[\left[\left[N_{k_l}^{(l)}P_N[\bar{\Phi}^{(S)}]D_{k_l}^{(l)}\right]_{k_l=1}^{d_l}\right]_{l=1}^{H}\right]$$

$$= \sum_{l=1}^{H}\sum_{k_l=1}^{d_l}P\left[N_{k_l}^{(l)}P_N[\bar{\Phi}^{(S)}]D_{k_l}^{(l)}\right].$$

From the construction of $N_{k_l}^{(l)}$, we have that for all $(l,k) \neq (l',k')$,

$$P[N_k^{(l)}P_N[\bar{\Phi}^{(S)}]D_k^{(l)}]P[N_{k'}^{(l')}P_N[\bar{\Phi}^{(S)}]D_{k'}^{(l')}] = 0.$$

Therefore,

$$\left\|P\left[P_N\left[\bar{\Phi}^{S}\right]D\right]\text{vec}(Y)\right\|_2^2 = \left\|\sum_{l=1}^{H}\sum_{k_l=1}^{d_l}P\left[N_{k_l}^{(l)}P_N[\bar{\Phi}^{(S)}]D_{k_l}^{(l)}\right]\text{vec}(Y)\right\|_2^2$$

$$= \sum_{l=1}^{H}\sum_{k_l=1}^{d_l}\left\|P\left[N_{k_l}^{(l)}P_N\left[\bar{\Phi}^{(S)}\right]D_{k_l}^{(l)}\right]\text{vec}(Y)\right\|_2^2.$$

$\square$

The following lemma plays a major role in the proof of theorem 2.

**Lemma 4.** *For any $t \in \{0, 1, \ldots, H\}$, every differentiable local minimum $\theta \in \Theta_{d_{H+1},t}$ satisfies that for any $l \in \{t, t+1, \ldots, H\}$,*

$$(\Phi^{(l)})^{\top}(\hat{Y}(X,\theta) - Y) = 0.$$

**Proof.** Fix $t$ to be a number in $\{0, 1, \ldots, H\}$. Let $\theta$ be a differentiable local minimum in $\Theta_{d_{H+1},t}$. Then, from the definition of a local minimum, there exists $\epsilon_1 > 0$ such that $L(\theta) \leq L(\theta')$ for all $\theta' \in B(\theta, \epsilon_1)$, and hence $L(\theta) \leq L(\theta')$ for all $\theta' \in \tilde{B}(\theta, \epsilon_1) \subseteq B(\theta, \epsilon_1)$, where $\tilde{B}(\theta, \epsilon_1) := B(\theta, \epsilon_1) \cap \{\theta \in \mathbb{R}^{d_\theta} : W^{(l+1)}(\theta)_{k',k} = 0 \text{ for all } l \in \{t+1, t+2, \cdots H-1\} \text{ and all } (k', k) \in S^{(l)} \times (\{1, \ldots, d_{l+1}\} \setminus S^{(l+1)})\}$ with the index sets $S^{(t+1)}, S^{(t+2)}, \ldots, S^{(H+1)}$ of the $\theta \in \Theta_{d_{H+1},t}$ in definition 1. Without loss of generality, we can permute the indices of the units within each layer such that for all $l \in \{t+1, t+2, \ldots, H+1\}$, $S^{(l)} \supseteq \{1, 2, \ldots, d_L\}$ with some $d_L \geq d_{H+1}$ in the definition of $\Theta_{d_{H+1},t}$ (see definition 1). Note that the considered activation functions $\sigma_{i,j}^{(l)}(z)$ are continuous and act linearly on $z \geq 0$. Thus, from the definition of $\Theta_{d_{H+1},t}$, there exists $\epsilon_2 > 0$ such that for all $\theta' \in \tilde{B}(\theta, \epsilon_2)$ and all $l \in \{t, t+1, \ldots, H\}$,

$$\hat{Y}(X, \theta') = \Phi^{(l)} \begin{bmatrix} A^{(l+1)} \\ C^{(l+1)} \end{bmatrix} A^{(l+2)} \cdots A^{(H+1)}$$
$$+ \sum_{l'=l}^{H-1} Z^{(l'+1)} C^{(l'+2)} A^{(l'+3)} \cdots A^{(H+1)}, \tag{A.1}$$

where $A^{(l)}$, $B^{(l)}$ and $C^{(l)}$ are submatrices of $W^{(l)}(\theta')$, and $Z^{(l)}$ is a submatrix of $\Phi^{(l)}(X, \theta')$ as defined below:

$$\begin{bmatrix} A^{(l)} & \xi^{(l)} \\ C^{(l)} & B^{(l)} \end{bmatrix} := W^{(l)}(\theta'),$$

and

$$Z^{(t+1)} := \sigma^{(t+1)}\left(\Phi^{(t)} \begin{bmatrix} \xi^{(t+1)} \\ B^{(t+1)} \end{bmatrix}\right) \text{ with } Z^{(l)} := \sigma^{(l)}(Z^{(l-1)}B^{(l)}) \text{ for } l \geq t+2.$$

Note that $Z^{(l)}$ depends only on $\Phi^{(t)}$, $\xi^{(t)}$, and $B^{(k)}$ for all $k \leq l$. Here, $\Phi^{(t)}$ does not depend on $A^{(l)}$ and $C^{(l)}$ for all $l \geq t+1$. That is, at each layer $l \in \{t+2, t+3, \ldots, H\}$, $A^{(l)} \in \mathbb{R}^{d_L \times d_L}$ connects $d_L$ linearly acting units to next $d_L$ linearly acting units, $B^{(l)} \in \mathbb{R}^{(d_{l-1}-d_L) \times (d_l-d_L)}$ connects other units to next other units (other units can include both nonlinear and linearly acting units), and $C^{(l)} \in \mathbb{R}^{(d_{l-1}-d_L) \times d_L}$ connects other units to next linearly acting units, with $d_L \geq d_{H+1}$. Here, $A^{(t+1)}$, $B^{(t+1)}$, $C^{(t+1)}$, and $\xi^{(t+1)}$ connect the possibly unstructured layer $\Phi^{(t)}$ to the next structured layer, $C^{(H+1)} \in \mathbb{R}^{(d_H-d_L) \times d_{H+1}}$ connects other units in the last hidden layer to the output units, and $A^{(H+1)} \in \mathbb{R}^{d_L \times d_{H+1}}$ connects linearly acting units in the last hidden layer to the output units.

Let $\epsilon_3 := \min(\epsilon_1, \epsilon_2)$. Let $l$ be an arbitrary fixed number in $\{t, t+1, \ldots, H\}$ in the following. Let $r := \hat{Y}(X, \theta) - Y$. Define

$$R^{(l+1)} := \begin{bmatrix} A^{(l+1)} \\ C^{(l+1)} \end{bmatrix}.$$

From the condition of differentiable local minimum, we have that

$$0 = \partial_{R^{(l+1)}} L(\theta) = \text{vec}((\Phi^{(l)})^\top r (A^{(l+2)} \cdots A^{(H+1)})^\top),$$

since otherwise, $R^{(l+1)}$ can be moved to the direction of $\partial_{R^{(l+1)}} L(\theta)$ with a sufficiently small magnitude $\epsilon_3' \in (0, \epsilon_3)$ and decrease the loss value. This implies that

$$(\Phi^{(l)})^\top r (A^{(l+2)} \cdots A^{(H+1)})^\top = 0.$$

If $\text{rank}(A^{(l+2)} \cdots A^{(H+1)}) \geq d_{H+1}$ or $l = H$, then this equation yields the desired statement of this lemma as $(\Phi^{(l)})^\top r = 0$. Hence, the rest of this proof considers the case of

$$\text{rank}((A^{(l+2)} \cdots A^{(H+1)})^\top) < d_{H+1} \text{ and } l \in \{t, t+1, \ldots, H-1\}.$$

Define an index $l^*$ as

$$l^* := \min\{l' \in \mathbb{Z}^+ : l+3 \leq l' \leq H+2 \ \wedge \ \text{rank}(A^{(l')} \cdots A^{(H+1)}) \geq d_{H+1}\},$$

where $A^{(H+2)} \cdots A^{(H+1)} := I_{d_{H+1}}$. This minimum exists since the set contains at least $H + 2$ (nonempty) and is finite. Then we have that $\text{rank}(A^{(l^*)} \cdots A^{(H+1)}) \geq d_{H+1}$ and $\text{rank}(A^{(l')} \cdots A^{(H+1)}) < d_{H+1}$ for all $l' \in \{l+2, l+3, \ldots, l^*-1\}$, since $\text{rank}(M_1 M_2) \leq \min(\text{rank}(M_1), \text{rank}(M_2))$. Therefore, for all $l' \in \{l+1, l+2, \ldots, l^*-2\}$, we have that $\text{Null}((A^{(l'+1)} \cdots A^{(H+1)})^\top) \neq 0$, and there exists a vector $u_{l'} \in \mathbb{R}^{d_L}$ such that

$$u_{l'} \in \text{Null}((A^{(l'+1)} \cdots A^{(H+1)})^\top) \text{ and } \|u_{l'}\|_2 = 1.$$

Let $u_{l'}$ denote such a vector for all $l' \in \{l+1, l+2, \ldots, l^*-2\}$. For all $l' \in \{l+2, l+3, \ldots, l^*-2\}$, define

$$\tilde{A}^{(l')}(v_{l'}) := A^{(l')} + v_{l'} u_{l'}^\top \text{ and } \tilde{R}^{(l+1)}(v_{l+1}) := R^{(l+1)} + v_{l+1} u_{l+1}^\top,$$

where $v_{l'} \in \mathbb{R}^{d_L}$ and $v_{l+1} \in \mathbb{R}^{d_l}$. Let $\tilde{\theta}(v_{l+1}, v_{l+2}, \ldots, v_{l^*-2})$ be $\theta$ with $A^{(l')}$ and $R^{(l+1)}$ being replaced by $\tilde{A}^{(l')}(v_{l'})$ and $\tilde{R}^{(l+1)}(v_{l+1})$ for all $l' \in \{l+2, l+3, \ldots, l^*-2\}$. Then for any $(v_{l+1}, v_{l+2}, \ldots, v_{l^*-2})$,

$$\hat{Y}(X, \tilde{\theta}(v_{l+1}, \ldots, v_{l^*-2})) = \hat{Y}(X, \theta) \text{ and } L(\tilde{\theta}(v_{l+1}, \ldots, v_{l^*-2})) = L(\theta),$$

since $\tilde{A}^{(l')}(v_{l'}) A^{(l'+1)} \cdots A^{(H+1)} = A^{(l')} A^{(l'+1)} \cdots A^{(H+1)}$ for all $l' \in \{l+2, l+3, \ldots, l^*-2\}$ and $\tilde{R}^{(l+1)}(v_{l+1}) A^{(l+2)} \cdots A^{(H+1)} = R^{(l+1)} A^{(l+2)} \cdots A^{(H+1)}$.

For any sufficiently small vector $(v_{l+1}, \ldots, v_{l^*-2})$ such that $\tilde{\theta}(v_{l+1}, \ldots, v_{l^*-2}) \in \bar{B}(\theta, \epsilon_3/2)$, if $\theta$ is a local minimum, every $\tilde{\theta}(v_{l+1}, \ldots, v_{l^*-2})$ is also a local minimum with respect to the entries of $A^{(l')}, B^{(l')}$, and $C^{(l')}$ for all $l'$ because there exists $\epsilon_3' = \epsilon_3/2 > 0$ such that

$$L(\tilde{\theta}(v_{l+1}, \ldots, v_{l^*-2})) = L(\theta) \leq L(\theta')$$

for all $\theta' \in \tilde{B}(\tilde{\theta}(v_{l+1}, \ldots, v_{l^*-2}), \epsilon_3') \subseteq \tilde{B}(\theta, \epsilon_3) \subseteq \tilde{B}(\theta, \epsilon_1) \subseteq B(\theta, \epsilon_1)$, where the first inclusion follows the triangle inequality. Thus, for any such $\tilde{\theta}(v_{l+1}, \ldots, v_{l^*-2})$ in the sufficiently small open ball, we have that

$$\partial_{A^{(l^*-1)}} L(\tilde{\theta}(v_{l+1}, \ldots, v_{l^*-2})) = 0,$$

where $\partial_{A^{(l^*-1)}} L(\tilde{\theta}(v_{l+1}, \ldots, v_{l^*-2}))$ exists within the sufficiently small open ball from equation A.1 (composed with the squared loss). In particular, by setting $v_{l+1} = 0$ and noticing that $\hat{Y}(X, \tilde{\theta}(v_{l+1}, \ldots, v_{l^*-2})) - Y = Y(X, \theta) - Y = r$,

$$
\begin{aligned}
0 &= \partial_{A^{(l^*-1)}} L(\tilde{\theta}(0, v_{l+2}, \ldots, v_{l^*-2})) \\
&= \left( \partial_{A^{(l^*-1)}} \hat{Y}(X, \tilde{\theta}(0, v_{l+2}, \ldots, v_{l^*-2})) \right)^\top \text{vec}(r),
\end{aligned}
$$

and hence

$$
\begin{aligned}
0 &= \partial_{A^{(l^*-1)}} L(\tilde{\theta}(v_{l+1}, \ldots, v_{l^*-2})) \\
&= \Big( \partial_{A^{(l^*-1)}} \Phi^{(l)}(v_{l+1} u_{l+1}^\top) \bar{A}^{(l+2)} \cdots \bar{A}^{(H+1)} \\
&\quad + \partial_{A^{(l^*-1)}} \hat{Y}(X, \tilde{\theta}(0, v_{l+2}, \ldots, v_{l^*-2})) \Big)^\top \text{vec}(r) \\
&= \left( \partial_{A^{(l^*-1)}} \Phi^{(l)}(v_{l+1} u_{l+1}^\top) \bar{A}^{(l+2)} \cdots \bar{A}^{(H+1)} \right)^\top \text{vec}(r),
\end{aligned}
$$

where

$$
\bar{A}^{(l')} = \begin{cases} \tilde{A}^{(l')}(v_{l'}) & \text{if } l' \in \{l+2, \ldots, l^*-2\} \\ A^{(l')} & \text{if } l' \notin \{l+2, \ldots, l^*-2\} \end{cases}.
$$

Since

$$
\begin{aligned}
&\partial_{A^{(l^*-1)}} \Phi^{(l)}(v_{l+1} u_{l+1}) \bar{A}^{(l+2)} \cdots \bar{A}^{(H+1)} \\
&= \left( (A^{(l^*)} \cdots A^{(H+1)})^\top \otimes \Phi^{(l)}(v_{l+1} u_{l+1}) \bar{A}^{(l+2)}(v_{l+2}) \cdots \tilde{A}^{(l^*-2)}(v_{l^*-2}) \right),
\end{aligned}
$$

this implies that

$$
A^{(l^*)} \cdots A^{(H+1)} r^\top \Phi^{(l)}(v_{l+1} u_{l+1}) \tilde{A}^{(l+2)}(v_{l+2}) \cdots \tilde{A}^{(l^*-2)}(v_{l^*-2}) = 0.
$$

By the definition of $l^*$, this implies that

$$
r^\top \Phi^{(l)}(v_{l+1} u_{l+1}) \tilde{A}^{(l+2)}(v_{l+2}) \cdots \tilde{A}^{(l^*-2)}(v_{l^*-2}) = 0,
$$

where $\tilde{A}^{(l+2)}(v_{l+2}) \cdots \tilde{A}^{(l+1)}(v_{l+1}) := I_{d_L}$.

We now show that for any sufficiently small vector $(v_{l+1}, \ldots, v_{l^*-2})$ such that $\tilde{\theta}(v_{l+1}, \ldots, v_{l^*-2}) \in \tilde{B}(\theta, \epsilon_3/2)$,

$$r^\top \Phi^{(l)}(v_{l+1}u_{l+1})\tilde{A}^{(l+2)}(v_{l+2}) \cdots \tilde{A}^{(j)}(v_j) = 0,$$

by induction on the index $j$ with the decreasing order $j = l^* - 2, l^* - 3, \ldots, l + 1$. The base case with $j = l^* - 2$ is proven above. Let $\tilde{A}^{(l)} := \tilde{A}^{(l)}(v_{l+2})$. For the inductive step, assuming that the statement holds for $j$, we show that it holds for $j - 1$ as

$$\begin{aligned}
0 &= r^\top \Phi^{(l)}(v_{l+1}u_{l+1}^\top)\tilde{A}^{(l+2)} \cdots \tilde{A}^{(j)} \\
&= r^\top \Phi^{(l)}(v_{l+1}u_{l+1}^\top)\tilde{A}^{(l+2)} \cdots A^{(j)} + r^\top \Phi^{(l)}(v_{l+1}u_{l+1}^\top)\tilde{A}^{(l+2)} \cdots \tilde{A}^{(j-1)}v_j u_j^\top \\
&= r^\top \Phi^{(l)}(v_{l+1}u_{l+1}^\top)\tilde{A}^{(l+2)} \cdots \tilde{A}^{(j-1)}v_j u_j^\top,
\end{aligned}$$

where the last line follows the fact that the first term in the second line is zero because of the inductive hypothesis with $v_j = 0$. Since $\|u_j\|_2 = 1$, by multiplying $u_j$ both sides from the right, we have that for any sufficiently small $v_j \in \mathbb{R}^{d_L}$,

$$r^\top \Phi^{(l)}(v_{l+1}u_{l+1}^\top)\tilde{A}^{(l+2)} \cdots \tilde{A}^{(j-1)}v_j = 0,$$

which implies that

$$r^\top \Phi^{(l)}(v_{l+1}u_{l+1}^\top)\tilde{A}^{(l+2)} \cdots \tilde{A}^{(j-1)} = 0.$$

This completes the inductive step and proves that

$$r^\top \Phi^{(l)}(v_{l+1}u_{l+1}^\top) = 0.$$

Since $\|u_{l+1}\|_2 = 1$, by multiplying $u_{l+1}$ both sides from the right, we have that for any sufficiently small $v_{l+1} \in \mathbb{R}^{d_l - d_L}$ such that $\tilde{\theta}(v_{l+1}, \ldots, v_{l^*-2}) \in \tilde{B}(\theta, \epsilon_3/2)$,

$$(\hat{Y}(X, \theta) - Y)^\top \Phi^{(l)}v_{l+1} = 0,$$

which implies that

$$(\Phi^{(l)})^\top(\hat{Y}(X, \theta) - Y) = 0.$$

$\square$

**A.1 Proof of Theorem 1.** From the first-order necessary condition of differentiable local minima,

$$0 = \partial_{W^{(H+1)}} L(\theta) = (D_1^{(H+1)})^\top \mathrm{vec}(\hat{Y}(X,\theta) - Y),$$

and $\hat{Y}(X,\theta) = D_1^{(H+1)} W^{(H+1)}$. From lemma 1 and the first-order necessary condition of differentiable local minima, $D^\top \mathrm{vec}(\hat{Y}(X,\theta) - Y) = 0$ and $\mathrm{vec}(\hat{Y}) = \frac{1}{H} D\theta_{1:H}$ where $\theta_{1:H} := \mathrm{vec}([W^{(l)}]_{l=1}^H)$. Combining these, we have that

$$\begin{bmatrix} D & D_1^{(H+1)} \end{bmatrix}^\top \left( \frac{1}{H+1} \begin{bmatrix} D & D_1^{(H+1)} \end{bmatrix} \theta - \mathrm{vec}(Y) \right) = 0,$$

where $\frac{1}{H+1} \begin{bmatrix} D & D_1^{(H+1)} \end{bmatrix} \theta = \mathrm{vec}(\hat{Y}(X,\theta))$. This implies that

$$\mathrm{vec}(\hat{Y}(X,\theta)) = P\left[\begin{bmatrix} D & D_1^{(H+1)} \end{bmatrix}\right] \mathrm{vec}(Y).$$

Therefore,

$$\begin{aligned} 2L(\theta) &= \|\mathrm{vec}(Y) - P\left[\begin{bmatrix} D & D_1^{(H+1)} \end{bmatrix}\right] \mathrm{vec}(Y)\|_2^2 \\ &= \|\mathrm{vec}(Y)\|_2^2 - \|P\left[\begin{bmatrix} D & D_1^{(H+1)} \end{bmatrix}\right] \mathrm{vec}(Y)\|_2^2, \end{aligned}$$

where the second line follows idempotence of the projection. Finally, decomposing the second term by directly following the proof of lemma 3 with $P_N\left[\bar{\Phi}^{(S)}\right] D$ being replaced by $\begin{bmatrix} D & D_1^{(H+1)} \end{bmatrix}$ yields the desired statement of this theorem. $\qquad \square$

**A.2 Proof of Theorem 2.** From lemma 4, we have that for any $l \in \{t, t+1, \ldots, H\}$,

$$(I_{d_{H+1}} \otimes \Phi^{(l)})^\top \mathrm{vec}(\hat{Y}(X,\theta) - Y) = 0. \tag{A.2}$$

From equation A.1, by noticing that $Z^{(l+1)} C^{(l+2)} = \Phi^{(l+1)} \begin{bmatrix} 0 & (C^{(l+2)})^\top \end{bmatrix}^\top$, we also have that

$$\mathrm{vec}(\hat{Y}(X,\theta)) = \sum_{l=t}^H (I_{d_{H+1}} \otimes \Phi^{(l)})^\top \mathrm{vec}(\bar{R}^{(l+1)} A^{(l+2)} \cdots A^{(H+1)}), \tag{A.3}$$

where $\bar{R}^{(t+1)} := \left[ (A^{(t+1)})^\top \quad (C^{(t+1)})^\top \right]^\top$ and $\bar{R}^{(l)} := \left[ 0 \quad (C^{(l)})^\top \right]^\top$ for $l \geq t + 2$. From lemma 1 and the first-order necessary condition of differentiable local minima, we also have that

$$D^\top \mathrm{vec}(\hat{Y}(X, \theta) - Y) = 0 \tag{A.4}$$

and

$$\mathrm{vec}(\hat{Y}) = \frac{1}{H} D\theta_{1:H}, \tag{A.5}$$

where $\theta_{1:H} := \mathrm{vec}([W^{(l)}]_{l=1}^H)$.

Combining equations A.2 to A.5 yields

$$\left[ \bar{\Phi} \quad D \right]^\top \left( \left[ \bar{\Phi} \quad D \right] \bar{\theta} - \mathrm{vec}(Y) \right) = 0,$$

where $\left[ \bar{\Phi} \quad D \right] \bar{\theta} = \mathrm{vec}(\hat{Y}(X, \theta))$, $\bar{\Phi} := [I_{d_{H+1}} \otimes \Phi^{(l)}]_{l=t}^H$, and

$$\bar{\theta} := \frac{1}{2} \left[ \ [\mathrm{vec}(\bar{R}^{(l+1)}A^{(l+2)} \cdots A^{(H+1)})^\top]_{l=t}^H \quad \frac{1}{H}\theta_{1:H}^\top \ \right]^\top.$$

This implies that

$$\mathrm{vec}(\hat{Y}(X, \theta)) = P\left[\left[ \bar{\Phi} \quad D \right]\right] \mathrm{vec}(Y).$$

Therefore, for any $S \subseteq (t, t+1, \ldots, H)$,

$$\begin{aligned}
2L(\theta) &= \|\mathrm{vec}(Y) - P\left[\left[ \bar{\Phi} \quad D \right]\right] \mathrm{vec}(Y)\|_2^2 \\
&\leq \|\mathrm{vec}(Y) - P\left[\left[ \bar{\Phi}^{(S)} \quad D \right]\right]\mathrm{vec}(Y)\|_2^2 \\
&= \|\mathrm{vec}(Y) - P[\bar{\Phi}^{(S)}]\mathrm{vec}(Y) - P[P_N[\bar{\Phi}^{(S)}]D]\mathrm{vec}(Y)\|_2^2 \\
&= \|P_N[\Phi^{(S)}]Y\|_F^2 - \|P[P_N[\bar{\Phi}^{(S)}]D]\mathrm{vec}(Y)\|_2^2, \tag{A.6}
\end{aligned}$$

where the second inequality holds because the column space of $\left[ \bar{\Phi} \quad D \right]$ includes the column space of $\left[ \bar{\Phi}^{(S)} \quad D \right]$. The third line follows lemma 2. The last line follows from the fact that $P_N[\bar{\Phi}^{(S)}] = (I - P[\bar{\Phi}^{(S)}])$ and

$$\begin{aligned}
\mathrm{vec}(Y)^\top P_N[\bar{\Phi}^{(S)}]^\top \, P[P_N[\bar{\Phi}^{(S)}]D]\mathrm{vec}(Y) &= \mathrm{vec}(Y)^\top P[P_N[\bar{\Phi}^{(S)}]D]\mathrm{vec}(Y) \\
&= \|P[P_N[\bar{\Phi}^{(S)}]D]\mathrm{vec}(Y)\|_2^2.
\end{aligned}$$

By applying lemma 3 to the second term on the right-hand side of equation A.6, we obtain the desired upper bound in theorem 2. Finally, we complete the proof by noticing that $\frac{1}{2}\|P_N[\Phi^{(S)}]Y\|_F^2$ is the global minimum value of basis function regression with the basis $\Phi^{(S)}$ for all $S \subseteq (0, 1, \ldots, H)$. This is because $\frac{1}{2}\|\Phi^{(S)}W - Y\|_F^2$ is convex in $W$ and, hence, $\partial_W \frac{1}{2}\|\Phi^{(S)}W - Y\|_F^2 = 0$ is a necessary and sufficient condition of global minima, solving which yields the global minimum value of $\frac{1}{2}\|P_N[\Phi^{(S)}]Y\|_F^2$. □

**A.3 Proof of Corollary 1.** The statement follows the proof of theorem 2 by noticing that lemma 4 still holds for the restriction of $L$ to $\mathcal{I}$ as $\tilde{B}(\theta, \epsilon_1) = B(\theta, \epsilon_1) \cap \mathcal{I}$, and by replacing $D_{k_l}^{(l)}$ by $\hat{D}_{k_l}^{(l)}$ in the proof, where $\hat{D}_{k_l}^{(l)}$ is obtained from the proof of lemma 1 by setting $W^{(l+1)}(\theta)_{k',k} = 0$ for $(k', k) \in S^{(l)} \times (\{1, \ldots, d_{l+1}\} \setminus S^{(l+1)})$ $(l = t+1, t+2, \cdots H-1)$ and by not considering their derivatives. □

**A.4 Proof of Corollary 2.** The statement follows the proof of theorem 2 by setting $C^{(l')} := 0$ for all $l' \notin \{t+1, H+1\}$ and setting $l \in \{t, H\}$ in the proof of lemma 4 (instead of $\{t, t+1, \ldots, H\}$). □

## Appendix B: Proofs for Probabilistic Statements

In the following lemma, we rewrite equation 3.2 in terms of the activation pattern, and data matrices $\begin{bmatrix} X & Y \end{bmatrix}$.

**Lemma 5.** *Every differentiable local minimizer $\theta$ of $L$ with the neural network 3.1 satisfies*

$$L(\theta) = \frac{1}{2}\|Y\|_2^2 - \frac{1}{2}\|P[\tilde{D}]Y\|_2^2, \tag{B.1}$$

*where*

$$\tilde{D} = \begin{bmatrix} \Lambda^{1,1}X & \Lambda^{1,2}X & \cdots & \Lambda^{1,d}X \end{bmatrix}. \tag{B.2}$$

**Proof.** With $r := \hat{Y}(X, \theta) - Y$, we have $L(\theta) = r^\top r/2$. For expression 3.1, we have

$$\hat{Y}(X, \theta) = \sum_{j=1}^d W_j^{(2)} \Lambda^{1,j} X W_{\cdot j}^{(1)}. \tag{B.3}$$

For any differentiable local minimum $\theta$, from the first-order condition,

$$0 = \partial_{W_{ij}^{(1)}} r^\top r/2 = W_j^{(2)} r^\top \Lambda^{1,j} X_{\cdot i}. \tag{B.4}$$

We conclude that if $W_j^{(2)} \neq 0$, then $r \perp \Lambda^{1,j} X_{\cdot i}$ for $1 \leq i \leq d_x$. In fact, we have the same conclusion even if $W_j^{(2)} = 0$. To prove it, we use the second-order condition as follows. We notice that if $W_j^{(2)} = 0$, then

$$
\begin{bmatrix} \partial^2_{W_{ij}^{(1)}} r^\top r/2 & \partial_{W_j^{(2)}} \partial_{W_{ij}^{(1)}} r^\top r/2 \\ \partial_{W_j^{(2)}} \partial_{W_{ij}^{(1)}} r^\top r/2 & \partial^2_{W_j^{(2)}} r^\top r/2 \end{bmatrix} = \begin{bmatrix} 0 & r^\top \Lambda^{1,j} X_{\cdot i} \\ r^\top \Lambda^{1,j} X_{\cdot i} & * \end{bmatrix}. \quad \text{(B.5)}
$$

By the second-order condition, the above matrix must be positive semidefinite, and we conclude that $r^\top \Lambda^{1,j} X_{\cdot i} = 0$. Therefore, $\hat{Y}(X, \theta) - Y$ is perpendicular to the column space of $\tilde{D}$. Moreover, from expression B.3, $\hat{Y}(X, \theta)$ is in the column space of $\tilde{D}$; $\hat{Y}(X, \theta)$ is the projection of $Y$ to the column space of $\tilde{D}$, $\hat{Y}(X, \theta) = P[\tilde{D}]Y$; and

$$
L(\theta) = \frac{1}{2} \|\hat{Y}(X, \theta) - Y\|_2^2 = \frac{1}{2} \|(I - P[\tilde{D}])Y\|_2^2 = \frac{1}{2} \|Y\|_2^2 - \frac{1}{2} \|P[\tilde{D}]Y\|_2^2.
$$
$$\text{(B.6)}$$

$\square$

From equation B.1, we expect that the larger the rank of the projection matrix $\tilde{D}$, the smaller is the loss $L(\theta)$. In the following lemma, we prove that under the conditions of the activation pattern matrix $\Lambda$. In the regime $d_x d \ll m$, we have rank $\tilde{D} = d_x d$. In the regime $d_x d \gg m$, we have rank $\tilde{D} = m$. As we show later, proposition 2 follows easily from the rank estimates of $\tilde{D}$.

**Lemma 6.** *Fix the activation pattern matrix $\Lambda := [\Lambda^k]_{k=1}^d \in \mathbb{R}^{m \times d}$. Let $X$ be a random $m \times d_x$ gaussian matrix, with each entry having mean zero and variance one. Then the matrix $\tilde{D}$ as defined in equation B.2 satisfies both of the following statements:*

    *i. If $m \geq 64 \ln^2(d_x dm/\delta^2) d_x d$ and $s_{\min}(\Lambda_I) \geq \delta$ for any index set $I \subseteq \{1, 2, \ldots, m\}$ with $|I| \geq m/2$, then rank $\hat{=} d_x d$ with probability at least $1 - e^{-m/(64 \ln(d_x dm/\delta^2))} - 2e^{-t}$.*

    *ii. If $dd_x \geq 2m \ln^2(md/\delta)$ with $d_x \geq \ln^2(dm)$ and $s_{\min}(\Lambda_I) \geq \delta$ for any index set $I \subseteq \{1, 2, \ldots, m\}$ with $|I| \leq d/2$, then rank $\tilde{D} = m$ with probability at least $1 - 2e^{-d_x/20}$.*

**Proof of Lemma 6.** We denote the event $\Omega_{\text{sum}}$ such that

$$
\Omega_{\text{sum}} = \{X : \|X\|_F^2 \leq 2md_x\}. \quad \text{(B.7)}
$$

Thanks to equation B.24 in lemma 7, $\mathbb{P}(\Omega_{\text{sum}}) \geq 1 - e^{-d_x m/8}$.

In the following, we first prove case i: that rank $\tilde{D} = d_x d$ with high probability. We identify the space $\mathbb{R}^{dd_x}$ with $d \times d_x$ matrix and fix $L = 2\lceil \ln(dm/\delta^2) \rceil$. We first prove that for any $V$ in the unit sphere in $\mathbb{R}^{dd_0}$, with probability at least $1 - e^{-m/(16L)}$, we have

$$\|\tilde{D}\text{vec}(V)\|_2^2 \geq \delta^2/(2L). \tag{B.8}$$

We notice that

$$\tilde{D}\text{vec}(V) = \sum_{j=1}^{d_x} \left( \sum_{i=1}^{d} \Lambda^i V_{ji} \right) X_{\cdot j} =: u.$$

Then $u$ is a gaussian vector in $\mathbb{R}^m$ with $k$th entry

$$u_k = \sum_{i=1}^{d_x} (\Lambda V)_{ki} X_{ki} \sim \mathcal{N}\left(0, a_k^2\right), \quad a_k^2 := \sum_{i=1}^{d_0} (\Lambda V)_{ki}^2.$$

Since by our assumption that the entries of $\Lambda$ are bounded by 1, we get

$$a_k^2 = \sum_{i=1}^{d_0} (\Lambda V)_{ki}^2 \leq \sum_{j=1}^{d} \Lambda_{kj}^2 \|V\|_F^2 \leq d.$$

We denote the sets $I_0 = \{1 \leq k \leq m : a_k^2 \leq \delta^2/m\}$ and

$$I_\ell = \{1 \leq k \leq m : e^{\ell-1}\delta^2/m < a_k^2 \leq e^\ell \delta^2/m\}, \quad 1 \leq \ell \leq \lceil \ln(dm/\delta^2) \rceil.$$

There are two cases: if there exists some $\ell \geq 1$ such that $|I_\ell| \geq m/L$, then thanks to equation B.25 in lemma 7, we have that with probability at least $1 - e^{-m/(16L)}$,

$$\|\Lambda(X)\text{vec}(V)\|_2^2 \geq \sum_{k \in I_\ell} u_k^2 \geq \frac{1}{2} \sum_{k \in I_\ell} a_k^2 \geq e^{\ell-1}\delta^2/(2L). \tag{B.9}$$

Otherwise, we have that $|I_0| \geq m(1 - \lceil \log_2(dm/\delta^2) \rceil/L) = m/2$. Then

$$\sum_{k \in I_0} a_k^2 \leq \sum_{k \in I_0} \delta^2/m < \delta^2.$$

However, by our assumption that $s_{\min}(\Lambda_{I_0}) \geq \delta$,

$$\sum_{k \in I_0} a_k^2 = \|\Lambda_{I_0} V\|_F^2 \geq s_{\min}^2(\Lambda_{I_0})\|V\|_F^2 \geq \delta^2.$$

This leads to a contradiction. Claim B.8 follows from claim B.9.

We take an $\varepsilon$-net of the unit sphere in $\mathbb{R}^{dd_x}$ and denote it by $\mathcal{E}$. The cardinality of the set $\mathcal{E}$ is at most $(2/\varepsilon)^{dd_x}$. We denote the event $\Omega$ such that the following holds:

$$\min_{V \in \mathcal{E}} \|\tilde{D}\mathrm{vec}(V)\|_2^2 \geq \delta^2/(2L). \tag{B.10}$$

Then by using a union bound, we get that the event $\Omega \cap \Omega_{\text{sum}}$ holds with probability at least $1 - e^{-m/(16L)}(2/\varepsilon)^{dd_x} - e^{-md_0/4}$.

Let $\hat{V}$ be a vector in the unit sphere of $\mathbb{R}^{dd_0}$. Then there exists a vector $V \in \mathcal{E}$ such that $\|V - \hat{V}\|_2 \leq \varepsilon$, and we have

$$\tilde{D}\mathrm{vec}(\hat{V}) = \tilde{D}\mathrm{vec}(V) + \tilde{D}\mathrm{vec}(\hat{V} - V). \tag{B.11}$$

From equations B.5 and B.10, for $X \in \Omega \cap \Omega_{\text{sum}}$, we have that

$$\left\|\tilde{D}\mathrm{vec}(V)\right\|_2^2 \geq \delta^2/(2L) \tag{B.12}$$

and

$$\begin{aligned}
\|\tilde{D}\mathrm{vec}(\hat{V} - V)\|_2^2 &\leq \sum_{k=1}^{m}\sum_{i=1}^{d_0}(\Lambda(\hat{V} - V))_{ki}^2\|x_k\|_2^2 \\
&\leq \sum_{k=1}^{m}\sum_{j=1}^{d}\Lambda_{kj}^2\|(\hat{V} - V)\|_F^2\|x_k\|_2^2 \leq \sum_{k=1}^{m}d\varepsilon^2\|x_k\|_2^2 \\
&\leq 2md_x d\varepsilon^2.
\end{aligned} \tag{B.13}$$

It follows from combining equations B.11 to B.13 that, we get that on the event $\Omega \cap \Omega_{\text{sum}}$,

$$\|\tilde{D}\mathrm{vec}(\hat{V})\|_2^2 \geq \delta^2/(4L),$$

provided that $\varepsilon \leq \delta/\sqrt{12d_x dmL}$. This implies that the smallest singular value of the matrix $\tilde{D}$ is at least $\delta^2/(4L)$, with probability

$$1 - e^{-m/(16L)}(2/\varepsilon)^{dd_x} - e^{-md_0/4} \geq 1 - e^{m/(32L)},$$

provided that $m \geq 32L\ln(d_x dm/\delta^2)d_x d$. This finishes the proof of Case i.

In the following we prove case ii that rank $\tilde{D} = m$ with high probability. We notice that for any vector $v \in \mathbb{R}^m$,

$$\|\tilde{D}^\top v\|_2^2 \leq \sum_{k=1}^m \sum_{i=1}^d \sum_{j=1}^{d_0} (v_k \Lambda_{ki} X_{kj})^2$$

$$\leq \sum_{j=1}^{d_0} \sum_{i=1}^d \left( \sum_{k=1}^m (v_k \Lambda_{ki})^2 \sum_{k=1}^m X_{kj}^2 \right) \qquad (B.14)$$

$$\leq d\|v\|_2^2 \|X\|_F^2.$$

In the event $\Omega_{\text{sum}}$ as defined in equation B.4, we have that $\|\tilde{D}^\top v\|_2^2 \leq 2dd_0 m\|v\|_2^2$ for any vector $v \in \mathbb{R}^m$.

In the following, we prove that for any vector $v \in \mathbb{R}^m$, if its $L$th largest entry (in absolute value) is at least $a$ for some $L \leq d/2$, then

$$\mathbb{P}(\|\tilde{D}^\top v\|_2^2 \geq a^2\delta^2 L d_0/2) \geq 1 - e^{-Ld_x/16}. \qquad (B.15)$$

We denote the vectors $u_i = [X_i^\top \Lambda^{1,1} v, X_i^\top \Lambda^{1,2} v, \ldots, X_i^\top \Lambda^{1,d} v]^\top$, for any $i = 1, 2, \ldots, d_x$. Then $\tilde{D}^\top v = [u_1, u_2, \ldots, u_{d_x}]^\top$. Moreover, $u_1, u_2, \ldots, u_{d_x} \in \mathbb{R}^d$ are independent and identically distributed (i.i.d) gaussian vectors, with mean zero and covariance matrix,

$$\Sigma = \Lambda^\top V^2 \Lambda,$$

where $V$ is the $m \times m$ diagonal matrix, with diagonal entries given by $v$. We denote the eigenvalues of $\Sigma$ as $\lambda_1(\Sigma) \geq \lambda_2(\Sigma) \geq \cdots \geq \lambda_d(\Sigma) \geq 0$. Then in distribution

$$\|u_i\|_2^2 = \lambda_1(\Sigma)z_{i1}^2 + \lambda_2(\Sigma)z_{i2}^2 + \cdots \lambda_d(\Sigma)z_{id}^2, \qquad (B.16)$$

where $\{z_{ij}\}_{1 \leq i \leq d_x, 1 \leq j \leq d}$ are independent gaussian random variables with mean zero and variance one. If the $L$th largest entry of $v$ (in absolute value) is at least $a$ for some $L \leq d/2$, we denote the index set $I = \{1 \leq k \leq m : |v_k| \geq a\}$, then

$$\Sigma = \Lambda^\top V^2 \Lambda \succ \Lambda^\top V_I^2 \Lambda \succ a^2 \Lambda_I^\top \Lambda_I.$$

Therefore, the $j$th largest eigenvalue of $\Sigma$ is at least the $j$th largest eigenvalue of $a^2 \Lambda_I^\top \Lambda_I$ for any $1 \leq j \leq d$. From our assumption, $s_{\min}(\Lambda_I) \geq \delta$, and the $L$th largest eigenvalue of $a^2 \Lambda_I^\top \Lambda_I$ is at least $a^2\delta^2$. Therefore, the $L$th largest eigenvalue of $\Sigma$ is at least $a^2\delta^2$, that is, $\lambda_L(\Sigma) \geq a^2\delta^2$. We can rewrite equation B.16 as

$$\|u_i\|_2^2 = \sum_{j=1}^{d} \lambda_j(\Sigma) z_{ij}^2 \geq a^2 \delta^2 \sum_{j=1}^{L} z_{ij}^2.$$

Thanks to equation B.25 in lemma 7,

$$\mathbb{P}\left(\|\tilde{D}^\top v\|_2^2 \geq a^2\delta^2 L d_0/2\right) = \mathbb{P}\left(\sum_{i=1}^{d_x} \|u_i\|_2^2 \geq a^2\delta^2 L d_0/2\right)$$

$$\geq \mathbb{P}\left(\sum_{i=1}^{d_x} a^2\delta^2 \sum_{j=1}^{L} z_{ij}^2 \geq a^2\delta^2 L d_0/2\right)$$

$$= \mathbb{P}\left(\sum_{i=1}^{d_x}\sum_{j=1}^{L} z_{ij}^2 \geq L d_0/2\right) \geq 1 - e^{-L d_0/16}.$$

This finishes the proof of claim B.15.

We take an $\varepsilon$-net of the unit sphere in $\mathbb{R}^m$ and denote it by $\mathcal{E}$. Let $\hat{v}$ be a vector in the unit sphere of $\mathbb{R}^m$; then there exists a vector $v \in \mathcal{E}$ such that $\|v - \hat{v}\|_2 \leq \varepsilon$, and we have

$$\tilde{D}^\top \hat{v} = \tilde{D}^\top v + \tilde{D}^\top(\hat{v} - v), \tag{B.17}$$

and in the event $\Omega_{\text{sum}}$ using equation D.14, we have

$$\|\tilde{D}^\top(\hat{v} - v)\|_2^2 \leq 2md_x d\varepsilon^2. \tag{B.18}$$

In the rest of the proof, we show that with high probability, $\|\tilde{D}^\top v\|_2^2$ is bounded away from zero for uniformly any $v \in \mathcal{E}$.

For any given vector $v$ in the unit sphere of $\mathbb{R}^m$, we sort its entries in absolute value:

$$|v_1^*| \geq |v_2^*| \geq \cdots \geq |v_m^*|.$$

We denote the sequence $1 = L_0 \leq L_1 \leq \cdots \leq L_p \leq L_{p+1} = m$, where $L_i = \lceil \ln^2(md/\delta) L_{i+1}/d_x \rceil$ for $1 \leq i \leq p$ and $L_1 \leq d_x/\ln^2(md/\delta)$. Then,

$$p = \lceil \ln m / \ln(d_x/\ln^2(md)) \rceil.$$

Thanks to our assumption that $dd_x \geq 2m\ln^2(md/\delta)$, we have $L_p \leq d/2$. We fix $\varepsilon$ as

$$\varepsilon := \frac{1}{2}\left(\frac{\delta}{4\sqrt{d}m}\right)^{p+1}. \tag{B.19}$$

We denote $q(v) = \min\{0 \le i \le p : |v_{L_i}^*| \ge 4\sqrt{d}m|v_{L_{i+1}+1}^*|/\delta\}$, where $v_{m+1}^* = 0$. We decompose the vector $v = v_1 + v_2$, where $v_1$ corresponds to the largest (in absolute value) $L_{q(v)+1}$ terms of $v$, and $v_2$ corresponds to the rest of terms of $v$. Letting $L = L_{q(v)}$ and $a = v_{L_{q(v)}}^*$ in equation B.15, we get

$$\mathbb{P}(\|\tilde{D}^\top v_1\|_2^2 \ge a^2\delta^2 L d_x/2) \ge 1 - e^{-Ld_0/16}. \tag{B.20}$$

By the definition of $q(v)$, we have

$$|a| = |v_{L_{q(v)}}^*| \ge \frac{\delta}{4\sqrt{d}m}|v_{L_{q(v)-1}}^*| \ge \cdots \ge \left(\frac{\delta}{4\sqrt{d}m}\right)^{q(v)}\frac{1}{\sqrt{m}} =: a_{q(v)}.$$

We denote the event $\Omega_q$, such that equation B.20 holds for any $v \in \mathcal{E}$ with $q(v) = q$. Since equation B.20 depends only on $L_{q+1}$ entries of $v$, by a union bound, we get

$$\mathbb{P}(\Omega_q) \ge 1 - e^{-L_q d_x/16}\binom{m}{L_{q+1}}(2/\varepsilon)^{L_{q+1}}$$

$$\ge 1 - e^{-L_q d_x/16 + L_{q+1}(\ln m + \ln(2/\varepsilon))} \ge 1 - e^{-L_q d_x/20}. \tag{B.21}$$

Moreover, $\|v_2\|_2^2 \le a^2\delta^2/(16dm)$, in the event $\Omega_{\text{sum}}$ using equation B.14, we have

$$\|\Lambda(X)^\top v_2\|_2^2 \le 2dd_x m\|v_2\|_2^2 \le a^2\delta^2 d_x/8. \tag{B.22}$$

It follows from combining equations B.20 and B.22, in the event $\Omega_q \cap \Omega_{\text{sum}}$ for any $v \in \mathcal{E}$ with $q(v) = q$, we get

$$\|\tilde{D}^\top v\|_2^2 \ge (\|\tilde{D}^\top v_1\| - \|\tilde{D}^\top v_2\|)^2 \ge a^2\delta^2 d_x/8 \ge a_q^2\delta^2 d_x/8. \tag{B.23}$$

In the event $\Omega_{\text{sum}} \cap_{q=0}^p \Omega_q$, it follows from combining equations B.17, B.18, and B.23 that

$$\|\tilde{D}^\top \hat{v}\|_2 \ge \|\tilde{D}^\top v\|_2^2 - \sqrt{2md_0d}\varepsilon \ge a_p\delta\sqrt{d_x/8} - \sqrt{2md_0d}\varepsilon \ge \left(\frac{\delta}{4\sqrt{d}m}\right)^{p+1}.$$

Moreover, thanks to equation B.21, $\Omega_{\text{sum}} \cap_{q=0}^p \Omega_q$ holds with probability at least

$$\mathbb{P}(\Omega_{\text{sum}} \cap_{q=0}^{p} \Omega_q) \geq 1 - \sum_{q=0}^{p} e^{-L_q d_x/20} - e^{-d_x m/8} \geq 1 - 2e^{-d_x/20}.$$

This finishes the proof of case ii. □

The following concentration inequalities for the square of gaussian random variables are from Laurent and Massart (2000).

**Lemma 7.** *Let the weights $0 \leq a_1, a_2, \ldots, a_n \leq K$, and $g_1, g_2, \ldots, g_n$ independent random gaussian variables with mean zero and variance one. Then the following inequalities hold for any positive $t$:*

$$\mathbb{P}\left(\sum_{i=1}^{n} a_i^2(g_i^2 - 1) \geq 2\sqrt{t}\left(\sum_{i=1}^{n} a_i^4\right)^{1/2} + 2K^2 t\right) \leq e^{-t}, \tag{B.24}$$

$$\mathbb{P}\left(\sum_{i=1}^{n} a_i^2(g_i^2 - 1) \leq -2\sqrt{t}\left(\sum_{i=1}^{n} a_i^4\right)^{1/2}\right) \leq e^{-t}. \tag{B.25}$$

**Proof of Proposition 2.** In case i from lemma 6, rank $\tilde{D} = d_x d$ with probability at least $1 - e^{-m/(64 \ln(d_x dm/\delta^2))}$. Since the statement immediately follows from theorem 1 if $\|Y\|_2 = 0$, we can focus on the case of $\|Y\|_2 \neq 0$. Conditioning on the event rank $\tilde{D} = d_x d$,

$$\frac{L(\theta)}{\|Y\|_2^2/2} = \frac{\|P_N[\tilde{D}]Y\|_2^2}{\|Y\|_2^2}. \tag{B.26}$$

The quantity in equation B.26 has the same law as

$$\frac{z_1^2 + z_2^2 + \cdots + z_{m-d_x d}^2}{z_1^2 + z_2^2 + \cdots + z_m^2},$$

where $z_1, z_2, \ldots, z_m$ are independent gaussian random variables with mean zero and variance one. From lemma 7, we get that with probability at least $1 - 2e^{-t}$,

$$\frac{z_1^2 + z_2^2 + \cdots + z_{m-d_x d}^2}{z_1^2 + z_2^2 + \cdots + z_m^2} \leq \left(1 + 6\sqrt{\frac{t}{m}}\right)\frac{m - d_x d}{m}. \tag{B.27}$$

Case i follows from combining equations B.26 and B.27.

In case ii, thanks to lemma 6, rank $\tilde{D} = m$ with probability at least $1 - 2e^{-d_x/20}$. Conditioning on the event rank $\tilde{D} = m$, we have $P[\tilde{D}]Y = Y$, and

$$L(\theta) = \frac{1}{2}\|Y\|_2^2 - \frac{1}{2}\|P[\tilde{D}]Y\|_2^2 = 0.$$

This finishes the proof of case ii. □

## Appendix C: Additional Experimental Details

By using the ground-truth network described in section 4.3, the synthetic data set was generated with i.i.d. random inputs $x$ and i.i.d. random weight matrices $W^{(l)}$. Each input $x$ was randomly sampled from the standard normal distribution, and each entry of the weight matrix $W^{(l)}$ was randomly sampled from a normal distribution with zero mean and normalized standard deviation ($\frac{2}{\sqrt{d_l}}$).

For training, we used a standard training procedure with mini-batch stochastic gradient decent (SGD) with momentum. The learning rate was set to 0.01. The momentum coefficient was set to 0.9 for the synthetic data set and 0.5 for the image data sets. The mini-batch size was set to 200 for the synthetic data set and 64 for the image data sets.

From the proof of theorem 1, $J(\theta) = \|(I - P[[D\ D_1^{(H+1)}]])\text{vec}(Y)\|_2^2$ for all $\theta$, which was used to numerically compute the values of $J(\theta)$. This is mainly because the form of $J(\theta)$ in theorem 1 may accumulate positive numerical errors for each $l \leq H$ and $k_l \leq d_l$ in the sum in its second term, which may easily cause a numerical overestimation of the effect of depth and width. To compute the projections, we adopted a method of computing a numerical cutoff criterion on singular values from Press, Teukolsky, Vetterling, and Flannery (2007) as (the numerical cutoff criterion) $= \frac{1}{2} \times$ (maximum singular value of $M$) $\times$ (machine precision of $M$) $\times$ ($\sqrt{d'+d+1}$), for a matrix of $M \in \mathbb{R}^{d' \times d}$. We also confirmed that the reported experimental results remained qualitatively unchanged with two other different cutoff criteria: a criterion based (Golub & Van Loan, 1996) as (the numerical cutoff criterion) $= \frac{1}{2}\|M\|_\infty \times$ (machine precision of $M$) (where $\|M\|_\infty = \max_{1 \leq i \leq d'} \sum_{j=1}^d |M_{i,j}|$ for a matrix of $M \in \mathbb{R}^{d' \times d}$), as well as another criterion based on Netlib Repository LAPACK documentation as (the numerical cutoff criterion) = (maximum singular value of $M$) $\times$ (machine precision of $M$).

## Acknowledgments

We gratefully acknowledge support from NSF grants 1523767 and 1723381; AFOSR grant FA9550-17-1-0165; ONR grant N00014-18-1-2847; Honda Research; and the MIT-Sensetime Alliance on AI. Any opinions, findings, and conclusions or recommendations expressed in this material are our own and do not necessarily reflect the views of our sponsors.

---